\documentclass[10pt,twocolumn,letterpaper]{article}

\usepackage[review]{cvpr}      
\definecolor{cvprblue}{rgb}{0.21,0.49,0.74}
\usepackage[pagebackref,breaklinks,colorlinks,allcolors=cvprblue]{hyperref}

\title{CodecMotion-OAD: Efficient Online Action Detection with Compressed-Domain Motion}

\author{First Author\\
Institution1\\
Institution1 address\\
{\tt\small firstauthor@i1.org}
\and
Second Author\\
Institution2\\
First line of institution2 address\\
{\tt\small secondauthor@i2.org}
}

\begin{document}
\maketitle
\begin{abstract}
Online action detection (OAD) recognizes ongoing actions in streaming video, often relying on RGB appearance and optical flow. Codec motion vectors (MVs) provide motion information at low computational cost, but directly replacing Flow with MVs can reduce detection accuracy. Our analysis shows that Flow and MVs capture similar dominant motion patterns; fine-grained motion details contribute less after RGB fusion, but motion still provides complementary cues that RGB alone cannot fully capture, helping maintain detection accuracy. The remaining gap between RGB+Flow and RGB+MV narrows substantially for conditional category identity but persists for action presence, where motion-only predictions provide complementary evidence. We propose CodecMotion-OAD to replace Flow with MVs in existing two-stream detectors. To our knowledge, this is the first use of codec MVs as a Flow substitute in frame-level causal OAD. Its unlabeled motion representation adaptation initializes an MV encoder with Flow-pretrained weights and adapts it through feature alignment. It also employs presence–identity routing to improve action presence estimation using complementary motion-only evidence while preserving the joint detector's conditional category distribution. On IPPP-re-encoded THUMOS'14, CodecMotion-OAD with MiniROAD achieves 69.60 mAP versus 70.39 for RGB+Flow, with 1.99× GPU and 11.13× CPU decoder-to-prediction speedups.
\end{abstract}    
\section{Introduction}
\label{sec:intro}

Online action detection (OAD) recognizes ongoing actions from streaming video \cite{degeest2016oad}. Despite advances in efficient temporal modeling \cite{xu2021lstr,zhao2022testra,pang2025cmert,an2023miniroad}, two-stream detectors still depend on dense optical flow (Flow), whose extraction can dominate processing cost. LSTR, for example, attributes approximately 90\% of runtime to motion-feature processing \cite{xu2021lstr}. Codec motion vectors (MVs), available during video decoding, offer a low-cost motion source without separate Flow estimation \cite{zhang2016emv,wu2018coviar}. Yet their sparse, block-wise, and reference-dependent nature makes direct substitution challenging. We investigate \textbf{whether and how codec MVs can effectively replace Flow in two-stream OAD}.

\begin{figure}[t]
  \centering

  \IfFileExists{figures/accuracy_latency.png}{%
    \includegraphics[width=\linewidth]{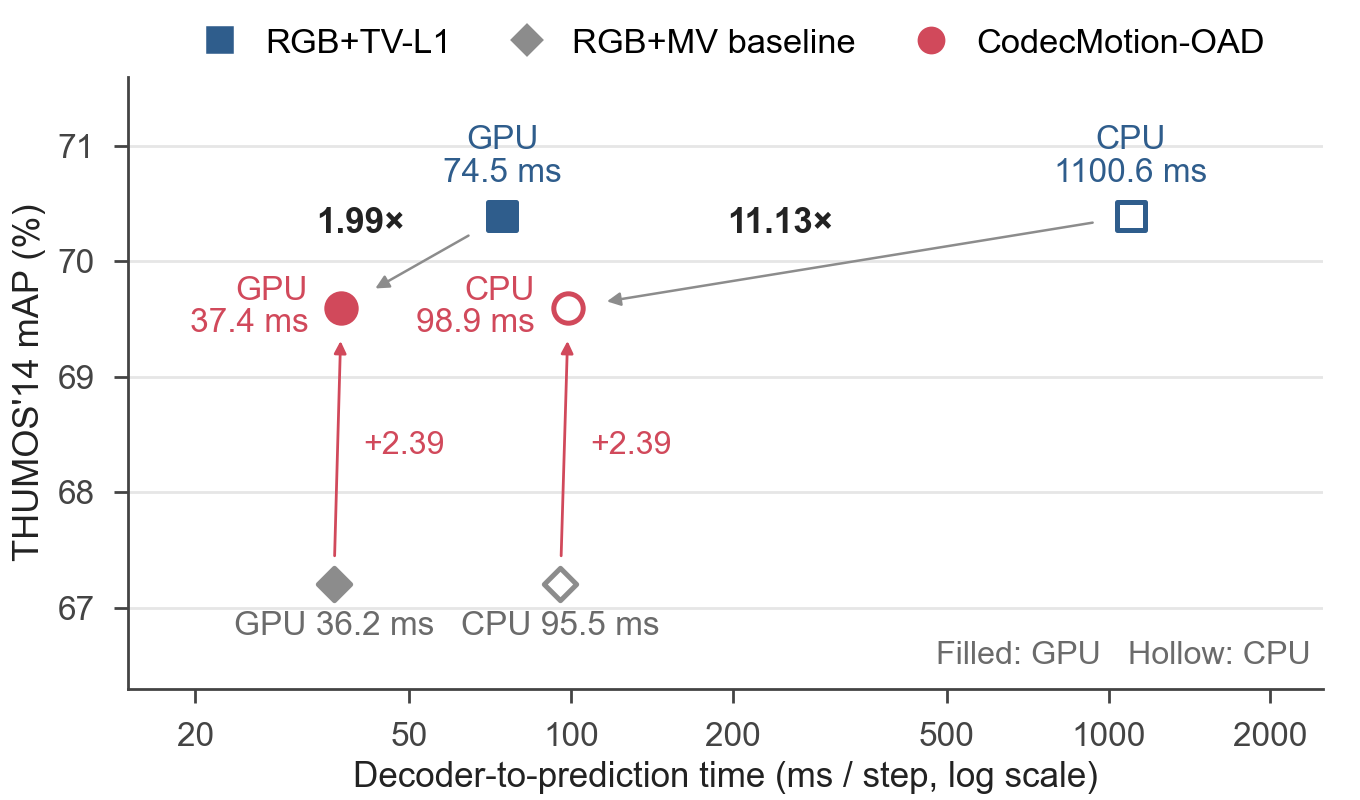}%
  }{%
    \fbox{\parbox[c][1.15in][c]{0.93\linewidth}{\centering Figure 1 image placeholder}}%
  }
  \caption{Accuracy--processing-time trade-off for MiniROAD on THUMOS'14, comparing RGB+Flow, adapted RGB+MV, and complete CodecMotion-OAD on GPU and CPU.}
  \label{fig:intro_tradeoff}
\end{figure}

We first ask whether fine-grained motion is essential once RGB appearance is available. Controlled Flow coarsening considerably degrades motion-only detection but causes much smaller losses in RGB+motion detection. Meanwhile, RGB+MV substantially outperforms RGB-only detection, even though MV-only is much weaker than Flow-only. These findings establish the complementary value of coarse motion and suggest partial functional overlap between RGB and fine-grained motion cues. We therefore retain the two-stream architecture while replacing its costly motion input.

Qualitative comparisons reveal similar dominant motion regions and directions between coarsened Flow and codec MVs. This similarity motivates reusing Flow-pretrained representations, while codec-specific sparsity and block structure call for adaptation. Our \textbf{Flow-to-MV representation adaptation} aligns an MV encoder with a frozen Flow encoder using paired motion inputs without action labels. The resulting representation can be reused across existing temporal detectors without separately optimizing motion encoders for each OAD architecture.

We further examine the roles of appearance and motion in \textbf{action presence} (action versus background) and \textbf{category identity} (which action is present). Adding RGB sharply narrows the Flow--MV gap in conditional identity, whereas the presence gap changes little. Moreover, MV-only and RGB+MV detectors exhibit similar presence performance but partly complementary presence errors. We therefore propose \textbf{presence--identity routing}, which combines the presence estimates of a joint RGB+MV detector and an MV-only auxiliary while preserving the joint detector's conditional category distribution.

Together, these components form \textbf{CodecMotion-OAD}. The auxiliary shares adapted MV features with the main detector, avoiding duplicate motion encoding. Experiments further show that a single lightweight MV-only MiniROAD auxiliary can be reused across four two-stream OAD architectures, eliminating the need for architecture-matched auxiliaries. The routing formulation also applies to Flow-based detection.

Experiments on THUMOS'14 demonstrate that CodecMotion-OAD substantially reduces decoder-to-prediction processing cost while maintaining competitive detection accuracy, as illustrated in Figure~\ref{fig:intro_tradeoff}. We use past-referenced IPPP coding to control decoding dependencies and additionally evaluate originally coded videos containing B pictures.

Our contributions are summarized as follows:

\begin{itemize}
  \item We analyze coarse motion and appearance--motion complementarity in two-stream OAD, distinguishing action presence from conditional category identity and identifying the distinct roles of RGB and motion predictions.
  \item We propose \textbf{CodecMotion-OAD}, combining unlabeled Flow-to-MV representation adaptation and presence--identity routing. Shared motion features and a reusable lightweight auxiliary enable integration with existing two-stream OAD detectors.
  \item We validate CodecMotion-OAD across four OAD architectures. On THUMOS'14 with MiniROAD, it achieves \textbf{69.60 mAP versus 70.39 for RGB+Flow}, while reducing decoder-to-prediction time from \textbf{74.5 to 37.4 ms/step on GPU (1.99$\times$)} and from \textbf{1100.6 to 98.9 ms/step on CPU (11.13$\times$)}. We further verify its effectiveness on originally coded videos containing B pictures.
\end{itemize}


\section{Related Work}
\label{sec:related}

\subsection{Online Action Detection}
\label{sec:related_oad}

Online action detection (OAD) recognizes ongoing actions from streaming video observations. Many established OAD detectors \cite{xu2019trn,eun2020idn} rely on two-stream RGB and optical-flow (Flow) features\cite{simonyan2014twostream} to capture complementary appearance and motion information. Representative, including LSTR \cite{xu2021lstr}, TeSTra \cite{zhao2022testra}, MiniROAD \cite{an2023miniroad}, and CMeRT \cite{pang2025cmert}, develop different mechanisms for temporal modeling through attention, memory, and recurrence. Beyond temporal detection, acquiring dense Flow features introduces substantial computational overhead. LSTR, for example, reports that motion-feature processing accounts for approximately 90\% of its runtime \cite{xu2021lstr}, highlighting the importance of motion acquisition in the overall OAD pipeline.

Several directions have been explored to improve processing efficiency. TeSTra investigates hardware-accelerated optical flow to reduce motion extraction latency \cite{zhao2022testra}. Other methods seek to avoid explicit Flow computation: GateHUB \cite{chen2022gatehub} and TLS-RWKV \cite{zhu2024tlsrwkv} explore Flow-free detection, E2E-LOAD learns video representations end-to-end \cite{cao2023e2eload}, and CAKE transfers motion knowledge into RGB-based representations \cite{hoang2026cake}. These approaches reduce reliance on conventional dense-flow pipelines through different combinations of feature learning, temporal modeling, and input design. However, these approaches may require specialized hardware or lead to considerable degradation in detection accuracy.

Our work investigates a complementary direction: \textbf{retaining an explicit motion stream while reducing its acquisition cost}. Rather than replacing the temporal detector or relying exclusively on RGB, we use codec motion vectors as a lightweight motion source. This enables existing two-stream OAD architectures to retain motion information without running a separate dense-flow estimator.

\begin{figure*}[t]
  \centering
  \IfFileExists{figures/visible.png}{%
    \includegraphics[width=\textwidth]{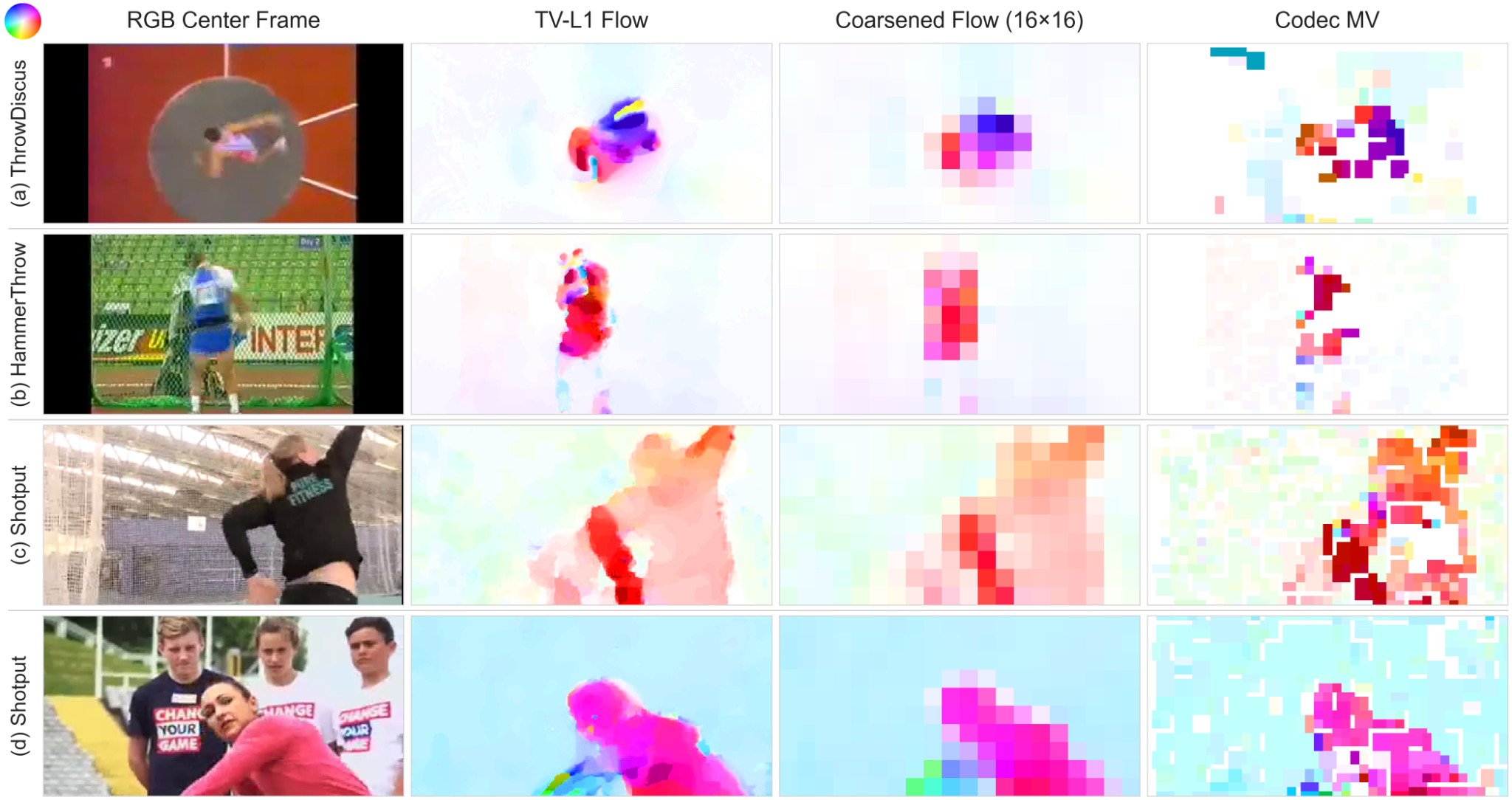}%
  }{%
    \fbox{\parbox[c][1.7in][c]{0.90\textwidth}{\centering Figure 2 image placeholder}}%
  }
  \caption{RGB frames, TV-L1 Flow, \(16\times16\) block-averaged Flow, and codec MVs on THUMOS'14 (ThrowDiscus, HammerThrow, and two Shotput examples). Motion visualizations within each row share direction and magnitude color scales; MVs additionally exhibit codec-dependent spatial structure.}
  \label{fig:motion_vis}
\end{figure*}

\subsection{Compressed-Domain Video Understanding}
\label{sec:related_compressed}

Compressed video contains motion vectors (MVs) and prediction residuals that can provide useful information for efficient video understanding \cite{yeo2008compressed}. Early work demonstrates the potential of these representations for action analysis. EMV-CNN transfers knowledge from optical flow to MVs for action recognition \cite{zhang2016emv}, while CoViAR \cite{wu2018coviar} and DMC-Net \cite{shou2019dmc} exploit compressed-domain information to improve the efficiency and effectiveness of video recognition. More recently, EdgeOAR investigates compressed-domain representations for online action recognition on edge devices \cite{luo2025edgeoar}. These studies establish codec-derived information as an alternative to computationally expensive motion estimation. However, its application to \textbf{frame-level causal OAD}, which requires continuous action predictions under temporal observation constraints, remains underexplored.

Unlike prior compressed-domain methods that often combine motion vectors and residuals\cite{chen2022mmvit,li2022meaci}, we focus on RGB and MVs. RGB can compensate for some discriminative information lost in coarse motion representations, while avoiding residual features reduces additional decoding and feature-computation overhead.




\section{Appearance and Motion in Online Action Detection}
\label{sec:analysis}

We investigate the feasibility of replacing dense optical flow (Flow) with codec motion vectors (MVs) in two-stream online action detection (OAD), and how to effectively exploit MVs. We first study the value of coarse motion and the effect of RGB fusion, then examine motion-structure similarity as a basis for representation reuse. Finally, we analyze action presence and category identity to explain the effect of RGB and motivate selective prediction fusion. Unless specified otherwise, experiments use MiniROAD on THUMOS'14 with a shared 4-fps, five-field feature interface.

\subsection{The Value of Coarse Motion in Two-Stream Detection}
\label{sec:analysis_coarse}

To isolate the importance of spatial detail, we average native \(320\times180\) TV-L1 Flow\cite{zach2007tvl1} within \(8\times8\), \(16\times16\), and \(32\times32\) blocks and tile the means back to the original resolution. This intervention removes fine detail while controlling for codec-dependent sparsity and reference structure. Each coarsened input uses a separately adapted Flow-initialized encoder (Section~\ref{sec:method_adaptation}). Table~\ref{tab:coarse_motion} compares motion-only and RGB-fused detection on 211 test videos.

\begin{table}[!t]
  \centering
  \caption{Coarsened Flow and adapted codec MV under the official MiniROAD configuration. Cosine measures similarity to original Flow features.}
  \label{tab:coarse_motion}
  \small
  \setlength{\tabcolsep}{4pt}

  \begin{tabular}{lrrr}
    \hline
    \textbf{Motion input} & \textbf{Cosine} & \textbf{Motion mAP} & \textbf{RGB+motion mAP} \\
    \hline
    Original Flow & 1.000 & 67.56 & 70.39 \\
    Flow \(8\times8\)  & 0.915 & 65.60 & 70.19 \\
    Flow \(16\times16\)& 0.860 & 57.27 & 68.83 \\
    Flow \(32\times32\) & 0.812 & 42.26 & 65.43 \\
    Codec MV & 0.828 & 52.52 & 67.21 \\
    \hline
  \end{tabular}

\end{table}

Two observations follow. \textbf{First, coarse motion remains complementary to RGB.} Even \(32\times32\) Flow yields 65.43 mAP with RGB, 5.47 points above RGB-only (59.96). Yet coarsening from original to \(32\times32\) Flow reduces motion-only mAP by 25.30 points and RGB+motion mAP by just 4.96 points. This contrast suggests that fine-motion cues and RGB have partially overlapping \emph{discriminative functions}. To identify where this functional overlap occurs, we further analyze action presence and category identity in Section~\ref{sec:analysis_presence}.

\textbf{Second, MVs are useful in a two-stream setting despite weaker standalone performance.} MV-only trails Flow-only by 15.04 points (52.52 versus 67.56), but the gap contracts to 3.18 points after RGB fusion (67.21 versus 70.39). RGB+MV also exceeds RGB-only by 7.25 points. We therefore retain RGB+motion detection, aiming to obtain an inexpensive motion stream that complements RGB and remains compatible with existing two-stream OAD detectors.

\subsection{Motion Structure and Flow Representation Reuse}
\label{sec:analysis_motion}

Figure~\ref{fig:motion_vis} compares dense Flow, \(16\times16\) coarsened Flow, and codec MVs. Coarsening removes local variations while preserving dominant motion regions and directions. MVs exhibit similar patterns in ThrowDiscus, HammerThrow, and Shotput, despite their codec-dependent block structures, sparsity, and reference-dependent displacements. Figure B1 provides additional examples, including missing MVs in I pictures.

These similarities motivate reusing Flow-pretrained representations, while differences in sparsity, block structure, and reference dependencies necessitate adaptation to MVs. We therefore adopt \textbf{task-agnostic Flow-to-MV representation adaptation}: initializing an MV encoder from Flow-pretrained weights and aligning its features with a frozen Flow encoder using paired motion inputs without action labels.

Unlabeled alignment avoids detector-specific supervised encoder adaptation and enables feature reuse across existing two-stream OAD architectures. The Flow encoder, without OAD-specific fine-tuning, provides a consistent reference for Flow--MV comparisons. Downstream detectors retain their supervised objectives, and Flow is needed only during offline adaptation, not inference. Section~\ref{sec:method_adaptation} details the adaptation objective, and Section~\ref{sec:exp_adaptation} evaluates its effectiveness.

\subsection{Action Presence and Category Identity}
\label{sec:analysis_presence}

Table~\ref{tab:coarse_motion} suggests functional overlap between RGB and fine-grained motion cues. We further investigate how this overlap affects two components of detection: \textbf{presence} (action versus background) and \textbf{identity} (action category given presence). For prediction distribution \(p_t\), background class \(\mathrm{bg}\), and action classes \(\mathcal C\), we define

\begin{equation}
  \begin{aligned}
  a_t &= 1-p_{t,\mathrm{bg}}, \\
  q_{t,c} &= \frac{p_{t,c}}{\sum_{k\in\mathcal C}p_{t,k}}.
\end{aligned}
  \label{eq:presence_identity}
\end{equation}

We report Presence AP and conditional Identity macro-AP on ground-truth action frames, together with AUROC and top-1 diagnostics; Ambiguous frames are excluded. Table~\ref{tab:presence_identity} compares Flow-minus-MV gaps with and without RGB, using MiniROAD and frame-wise probes to examine the effect across different temporal modeling settings.

\begin{table*}[!t]
  \centering
  \caption{Flow-minus-MV performance gaps in percentage points; positive values favor Flow. Identity is evaluated on ground-truth action frames.}
  \label{tab:presence_identity}
  \small
  \setlength{\tabcolsep}{4pt}
  \begin{tabular*}{\textwidth}{@{\extracolsep{\fill}}lrrrr@{}}
    \hline
    \textbf{Metric} & \textbf{Probe: motion} & \textbf{Probe: +RGB} & \textbf{MiniROAD: motion} & \textbf{MiniROAD: +RGB} \\
    \hline
    mAP & +8.8 & +2.5 & +15.04 & +3.18 \\
    Presence AP & +5.3 & +3.2 & +3.62 & +3.10 \\
    Presence AUROC & +3.1 & +1.7 & +2.51 & +1.69 \\
    Identity macro-AP & +10.4 & +1.0 & +15.17 & +1.00 \\
    Identity top-1 & +10.6 & +1.6 & +9.55 & +3.13 \\
    \hline
  \end{tabular*}
\end{table*}

With MiniROAD, RGB reduces the Identity macro-AP gap from 15.17 to 1.00 points, while the Presence AP gap changes only from 3.62 to 3.10. Frame-wise probes exhibit the same asymmetry. In absolute terms, adding RGB to MV-only detection raises Identity macro-AP from 75.97 to 95.52, while Presence AP changes from 79.56 to 79.69. For Flow-only detection, identity increases from 91.14 to 96.52, while presence changes from 83.18 to 82.79.

These results clarify the functional overlap observed in Table~\ref{tab:coarse_motion}. RGB substantially supplements \emph{category discrimination}, reducing the importance of fine-grained motion cues for conditional identity. In contrast, RGB provides limited additional presence discrimination once motion is available, leaving a relatively persistent Flow--MV presence gap. Thus, action presence emerges as a key remaining limitation when substituting MVs for Flow.

\begin{figure}[t]
  \centering
  \IfFileExists{figures/flow_mv_gap.png}{%
    \includegraphics[width=\linewidth]{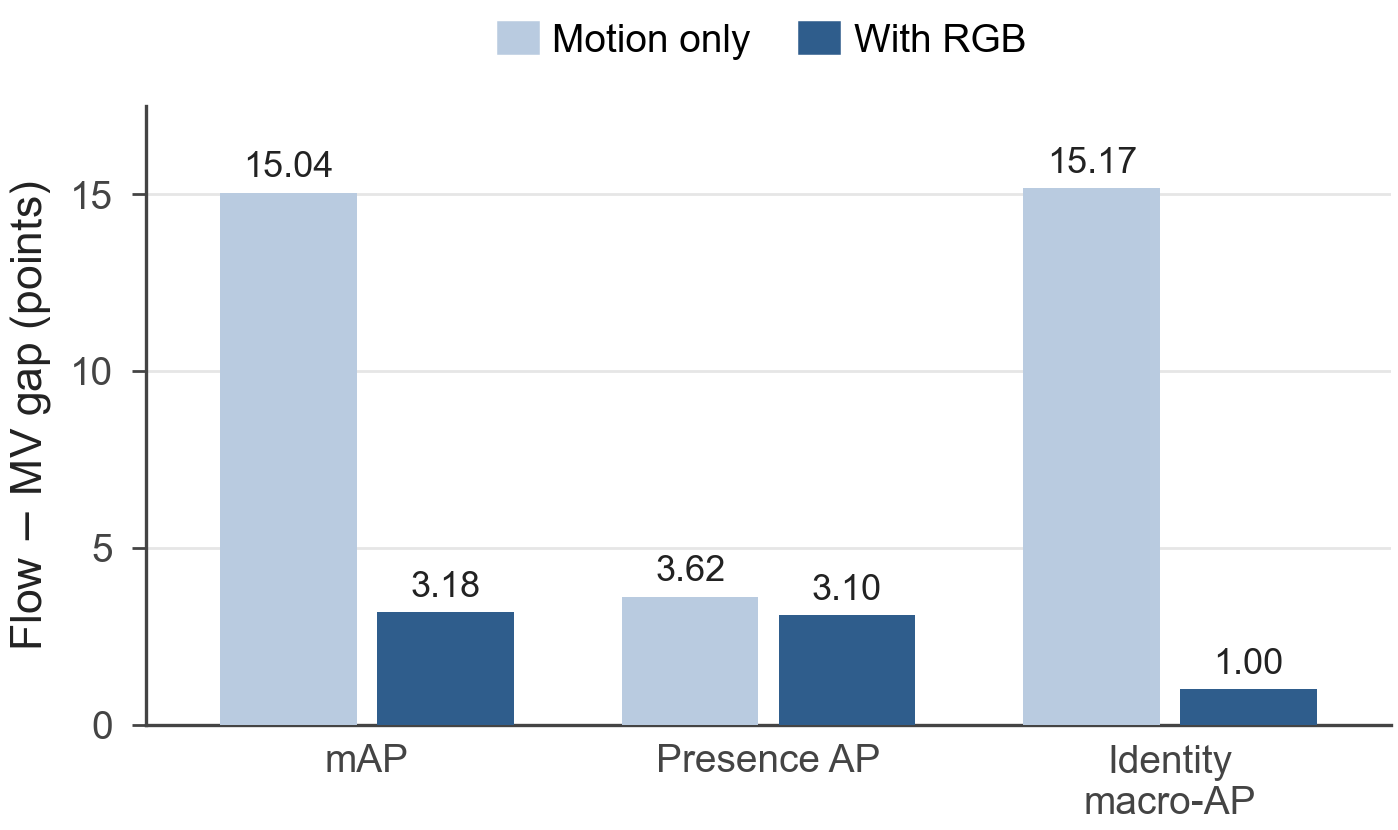}%
  }{%
    \fbox{\parbox[c][1.2in][c]{0.90\linewidth}{\centering Figure 3 image placeholder}}%
  }
  \caption{Flow-minus-MV gaps for MiniROAD before and after RGB fusion: identity and overall detection gaps contract markedly, whereas the presence gap changes little.}
  \label{fig:presence_gaps}
\end{figure}

Feature diagnostics further distinguish representational similarity from downstream task utility. Approximately 59\% of Flow-feature variance remains unexplained by MVs, and RGB poorly predicts that residual (\(R^2=0.026\)). Yet adding the residual improves MV-only probe mAP by 8.4 points but RGB+MV by only 1.9 (Appendix A.3). This further supports the reduced conditional task value of fine-grained motion information after appearance fusion.

The remaining presence gap motivates examining complementary motion-only predictions. MV-only and RGB+MV have similar Presence AP (79.56 versus 79.69) but substantially different Identity macro-AP (75.97 versus 95.52). Their false-positive sets overlap by only 30.2\% Jaccard at threshold 0.5 (Appendix A.4), indicating partly complementary presence errors. These observations suggest that MV-only predictions can provide additional presence evidence, while the joint RGB+MV detector supplies stronger conditional category discrimination.

We therefore propose \textbf{presence--identity routing}, which combines the two presence estimates while retaining the RGB+motion detector's conditional category distribution. The MV-only auxiliary shares adapted MV features with the main detector, avoiding duplicate encoding. The routing principle also extends to Flow-based detection. Section~\ref{sec:method} defines the method, and Section~\ref{sec:experiments} evaluates its detection gains and computational cost.




\section{Method}
\label{sec:method}

\begin{figure*}[t]
  \centering
  \IfFileExists{figures/overview.png}{%
    \includegraphics[width=\textwidth]{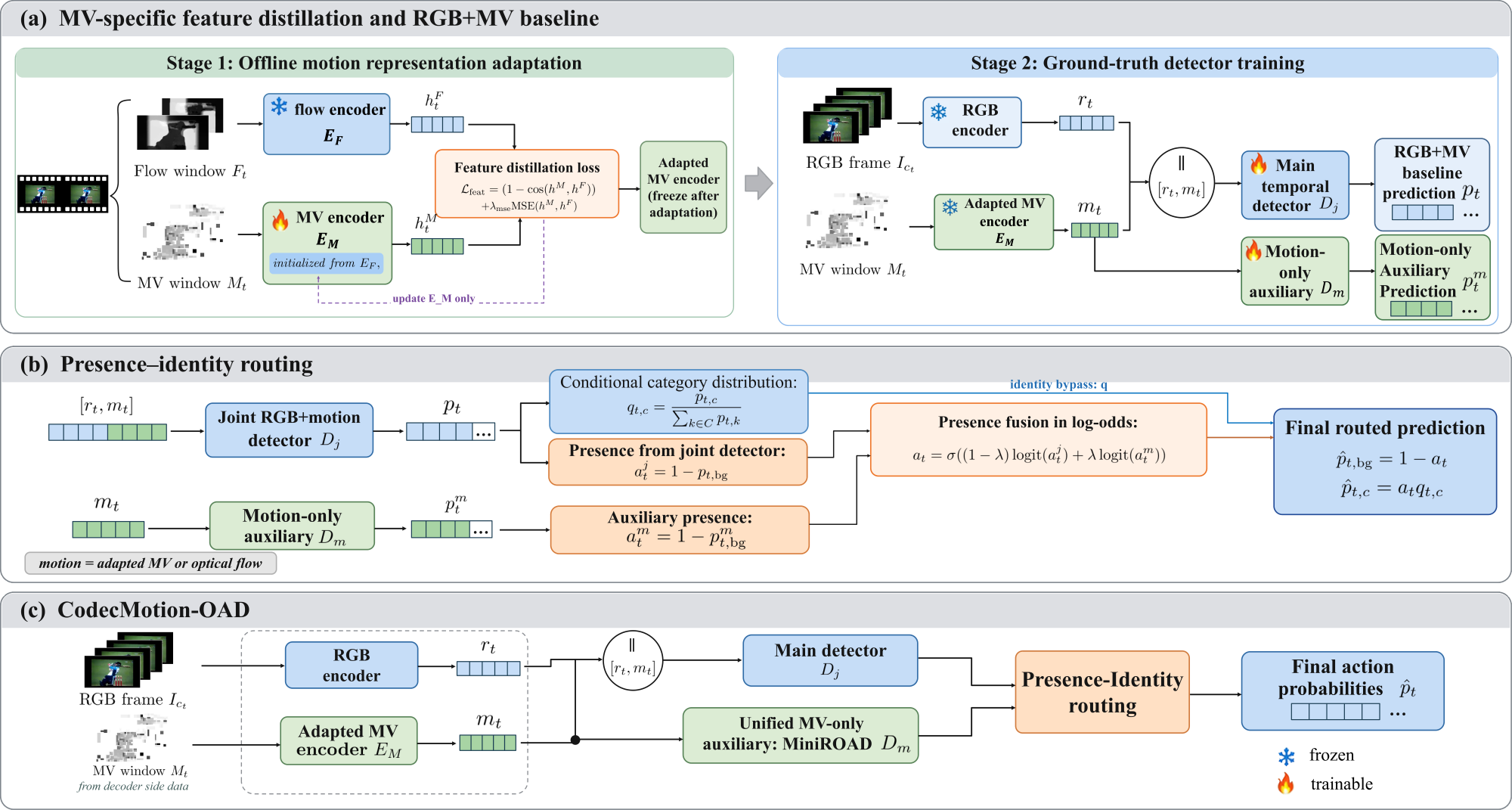}%
  }{%
    \fbox{\parbox[c][1.7in][c]{0.90\textwidth}{\centering Figure 4 image placeholder}}%
  }
  \caption{Overview of CodecMotion-OAD. (a) Offline Flow-to-MV representation adaptation followed by supervised training of joint and motion-only detectors. (b) Presence--identity routing combines presence estimates from a joint RGB+motion detector and a motion-only auxiliary while retaining the joint conditional category distribution. The routing rule supports either MVs or Flow. (c) The complete CodecMotion-OAD system uses shared adapted MV features and a lightweight MV-only MiniROAD auxiliary.}
  \label{fig:overview}
\end{figure*}

CodecMotion-OAD consists of two components (Figure~\ref{fig:overview}): Flow-to-MV representation adaptation and presence--identity routing. The former adapts a Flow-pretrained encoder to codec motion vectors (MVs), providing reusable motion features for existing two-stream OAD detectors. The latter combines action-presence estimates from a joint RGB+motion detector and a motion-only auxiliary while retaining the joint detector's conditional category distribution. The temporal detectors and their supervised objectives remain unchanged.

\subsection{Codec Motion Extraction and Online Availability}
\label{sec:method_extraction}

We follow the standard OAD feature-extraction protocol for both Flow and codec MVs. At feature step \(t\), five consecutive motion fields ending at \(e_t=c_t+2\) are paired with the RGB center frame \(I_{c_t}=I_{e_t-2}\). Each field contains horizontal and vertical displacement components, forming a ten-channel motion input encoded by ResNet-50 \cite{he2016resnet} into a 2048-dimensional feature. RGB, Flow, and MV features follow the same nominal 4-fps grid, with sampling and alignment details provided in Appendix C.2.

Codec MVs are extracted from decoder motion-vector data, converted to pixel displacements, and rasterized over their destination blocks. Regions without valid motion vectors are filled with zeros. The resulting fields are resized and cached before feature encoding, providing the same input interface as Flow.

For online detection, each feature step is processed once its required source observations become available. Unlike Flow, codec MV availability additionally depends on inter-picture prediction. Let \(\mathcal W_t\) denote the five motion-field indices and \(\mathcal R^*(i)\) the set of source pictures required to decode picture \(i\), including the picture itself and its recursive references. The earliest source-frame index at which the MV input becomes available is

\begin{equation}
  \tau_t^M=\max_{i\in\mathcal W_t}\max_{r\in\mathcal R^*(i)}r.
  \label{eq:mv_availability}
\end{equation}

Under past-referenced IPPP coding, all references precede their coded pictures in display order, yielding \(\tau_t^M\le e_t\). In contrast, B pictures may reference later pictures\cite{wiegand2003h264}, extending decoding dependencies beyond the selected motion window. We therefore use IPPP-re-encoded videos for controlled comparisons. To evaluate original coded streams containing B pictures, we retain only past-directed MVs to minimize dependence on future observations (Section~\ref{sec:exp_original}).

Once the required observations are available, completed feature steps are processed sequentially without accessing future feature steps. Appendix C.1--C.3 provides the coding configurations, reference-dependency analysis, and source-frame availability measurements.

\subsection{Flow-to-MV Representation Adaptation}
\label{sec:method_adaptation}

We initialize the MV encoder \(E_M\) from a Flow-pretrained ResNet-50 and adapt it through unlabeled feature alignment. A frozen Flow encoder \(E_F\) provides representation targets from paired TV-L1 Flow inputs.

For temporally aligned MV and Flow inputs \(M_t\) and \(F_t\), their encoded features are

\begin{equation}
  h_t^M=E_M(M_t),\qquad h_t^F=E_F(F_t).
  \label{eq:paired_features}
\end{equation}

We optimize the MV encoder using a combination of cosine alignment and mean-squared feature matching:

\begin{equation}
  \begin{aligned}\mathcal{L}_{\mathrm{feat}} & = \left(1-\frac{\langle h_t^M,h_t^F\rangle}{\|h_t^M\|_2\|h_t^F\|_2}\right) \\ &\quad +\frac{\lambda_{\mathrm{mse}}}{d}\|h_t^M-h_t^F\|_2^2,\end{aligned}
  \label{eq:feat_align}
\end{equation}

where \(d=2048\) and \(\lambda_{\mathrm{mse}}=0.25\). The cosine term aligns feature directions, while the MSE term matches feature values.

Only \(E_M\) is updated during adaptation; the Flow teacher and BatchNorm statistics remain frozen. Adaptation uses paired motion inputs from training videos without action labels. The adapted encoder is subsequently frozen and used to extract MV features for downstream OAD detectors, which retain their original supervised training objectives.

This task-agnostic procedure produces a common MV representation reusable across different two-stream temporal detectors. During deployment, the adapted MV encoder operates independently, without Flow estimation or the teacher encoder.

\begin{table*}[!t]
  \centering
  \caption{THUMOS'14 detection mAP.
  Adapted MV excludes routing; Unified MiniROAD
  denotes the complete system.}
  \label{tab:main}
  \small
  \setlength{\tabcolsep}{8pt}
  \begin{tabular*}{\textwidth}{@{\extracolsep{\fill}}lrrrr@{}}
    \hline
    \textbf{Detector} &
    \textbf{RGB+Flow} &
    \textbf{RGB+MV (Adapted)} &
    \textbf{Matched routing} &
    \textbf{Unified MiniROAD} \\
    \hline
    MiniROAD & 70.39 & 67.21 & 69.60 & 69.60 \\
    CMeRT    & 70.50 & 68.31 & 69.80 & 70.53 \\
    LSTR     & 69.60 & 66.74 & 67.61 & 69.76 \\
    TeSTra   & 70.50 & 68.20 & 69.60 & 71.11 \\
    \hline
  \end{tabular*}
\end{table*}

\subsection{Presence--Identity Routing}
\label{sec:method_routing}

We employ a joint RGB+motion detector and a motion-only auxiliary to obtain complementary predictions. In CodecMotion-OAD, both detectors share the frozen adapted MV encoder, avoiding duplicate motion encoding.

Let \(r_t\) and \(m_t\) denote the RGB and motion features at step \(t\). The joint detector \(D_j\) and motion-only auxiliary \(D_m\) produce

\begin{equation}
  \begin{aligned} p_t &= D_j([r_t,m_t];s_{t-1}), \\ p_t^m &= D_m(m_t;s_{t-1}^m),\end{aligned}
  \label{eq:joint_aux}
\end{equation}

where \(s_{t-1}\) and \(s_{t-1}^m\) denote their independent temporal states. Both detectors are trained separately using ground-truth action supervision. We consider architecture-matched motion-only auxiliaries and a unified lightweight MiniROAD auxiliary, which is reused across four main-detector architectures. The same routing formulation is also evaluated with Flow-based detectors and Flow-only auxiliaries.

Rather than averaging the complete prediction distributions, we separate action presence from conditional category identity. For action classes \(\mathcal C\) and background class \(\mathrm{bg}\), the joint and auxiliary presence estimates are

\begin{equation}
  a_t^j=1-p_{t,\mathrm{bg}},\qquad a_t^m=1-p_{t,\mathrm{bg}}^m.
  \label{eq:presence_estimates}
\end{equation}

The joint detector provides the conditional category distribution

\begin{equation}
  q_{t,c}=\frac{p_{t,c}}{\sum_{k\in\mathcal C}p_{t,k}},\qquad c\in\mathcal C.
  \label{eq:conditional_identity}
\end{equation}

We interpolate the log-odds of the two presence estimates with auxiliary weight \(\lambda\in[0,1]\):

\begin{equation}
  \begin{aligned} u_t^j &= \log\frac{a_t^j}{1-a_t^j}, \\ u_t^m &= \log\frac{a_t^m}{1-a_t^m},\end{aligned}
  \label{eq:presence_logodds}
\end{equation}

\begin{equation}
  a_t=\sigma\left((1-\lambda)u_t^j+\lambda u_t^m\right),
  \label{eq:presence_fusion}
\end{equation}

where \(\sigma\) denotes the sigmoid function.

The final prediction combines the fused presence estimate with the joint detector's conditional category distribution:

\begin{equation}
  \boxed{\begin{aligned}\widehat p_{t,\mathrm{bg}} &=1-a_t, \\ \widehat p_{t,c} &=a_tq_{t,c},\quad c\in\mathcal C.\end{aligned}}
  \label{eq:routing_output}
\end{equation}

Routing adjusts action presence while preserving \(q_{t,c}\), leaving conditional Identity macro-AP unchanged. The resulting action probabilities are rescaled by the fused presence estimate. Fusion introduces no trainable parameters. We fix \(\lambda=0.5\) for MV-based experiments and evaluate weight sensitivity in Section~\ref{sec:exp_aux} and Appendix A.6.

For numerical stability, presence probabilities are clipped to \([10^{-6},1-10^{-6}]\), and the denominator of \(q_{t,c}\) is bounded below by \(10^{-12}\). On THUMOS'14, the final 22-channel output assigns zero probability to Ambiguous, whose labeled frames are excluded from evaluation.

During inference, each completed feature step is encoded once and passed to both detectors. Their predictions are then combined through presence--identity routing. Only the auxiliary temporal update and scalar fusion add computation beyond the main RGB+MV detector, as quantified in Section~\ref{sec:exp_efficiency}.


\section{Experiments}
\label{sec:experiments}

We evaluate CodecMotion-OAD across four detectors, then examine processing cost, representation adaptation, fusion, auxiliary modality, and original coded streams.

\subsection{Experimental Setup}
\label{sec:exp_setup}

We evaluate on THUMOS'14 \cite{idrees2017thumos}, using 200 training and 211 test videos across 20 action classes. Frame-level mAP is the primary metric, excluding Ambiguous frames. We evaluate MiniROAD, CMeRT, LSTR, and TeSTra under a unified protocol, using Kinetics-pretrained\cite{carreira2017kinetics} ResNet-50 feature encoders for RGB and motion inputs. RGB, Flow, and MV features are aligned on a nominal 4-fps grid and extracted from the same IPPP-re-encoded videos. The adapted MV encoder is frozen during supervised detector training, while each temporal detector retains its original training objective. Processing efficiency is measured on an NVIDIA V100 GPU and an Intel Xeon Gold 6240 CPU. Additional training, feature-extraction, and timing details are provided in Appendices C and D.

\subsection{Detection Performance}
\label{sec:exp_results}

Table~\ref{tab:main} compares adapted RGB+MV detectors and their routing variants across four temporal architectures. We first pair each joint detector with an MV-only auxiliary of the same architecture to assess matched routing.
\begin{table}[!tbp]
  \centering
  \caption{Decoder-to-prediction latency (ms/step),
  including auxiliary execution and routing.
  RGB+MV denotes adapted codec motion vectors.}
  \label{tab:efficiency}
  \small
  \setlength{\tabcolsep}{3pt}
  \begin{tabular}{@{}llrrr@{}}
    \hline
    \textbf{Detector} & \textbf{System} &
    \textbf{mAP} & \textbf{GPU} & \textbf{CPU} \\
    \hline
    MiniROAD & RGB+Flow  & 70.39 & 74.5 & 1100.6 \\
             & RGB+MV    & 67.21 & 36.2 & 95.5 \\
             & CodecMotion-OAD & 69.60 & 37.4 & 98.9 \\
    \hline
    CMeRT    & RGB+Flow  & 70.50 & 94.4 & 1153.2 \\
             & RGB+MV    & 68.31 & 58.9 & 144.9 \\
             & CodecMotion-OAD & 70.53 & 59.7 & 149.5 \\
    \hline
  \end{tabular}
\end{table}

\begin{table}[!tbp]
  \centering
  \caption{CMeRT adaptation ablation; cosine similarity to the TV-L1 teacher.}
  \label{tab:adapt}
  \small
  \setlength{\tabcolsep}{4pt}
  \begin{tabular}{@{}lrr@{}}
    \hline
    \textbf{MV encoder} & \textbf{Cosine} & \textbf{RGB+MV mAP} \\
    \hline
    Fixed Flow-pretrained & 0.456 & 63.52 \\
    Adapted & 0.828 & 68.31 \\
    \hline
  \end{tabular}
\end{table}

\begin{table*}[!t]
  \centering
  \caption{Fusion comparison (mAP; \(\lambda=0.5\)).}
  \label{tab:fusion}
  \small
  \setlength{\tabcolsep}{4pt}
  \begin{tabular}{@{}llrrrr@{}}
    \hline
    \textbf{Main} & \textbf{MV-only auxiliary} & \textbf{RGB+MV (Adapted)} & \textbf{Probability avg.} & \textbf{Arithmetic presence} & \textbf{Log-odds (ours)} \\
    \hline
    MiniROAD & MiniROAD & 67.21 & 66.59 & 69.23 & 69.60 \\
    CMeRT & CMeRT & 68.31 & 67.99 & 69.50 & 69.80 \\
    CMeRT & MiniROAD & 68.31 & 68.91 & 70.14 & 70.53 \\
    LSTR & LSTR & 66.74 & 63.46 & 66.86 & 67.61 \\
    LSTR & MiniROAD & 66.74 & 67.47 & 68.82 & 69.76 \\
    TeSTra & TeSTra & 68.20 & 66.01 & 69.27 & 69.60 \\
    TeSTra & MiniROAD & 68.20 & 68.03 & 70.39 & 71.11 \\
    \hline
  \end{tabular}
\end{table*}

Matched routing improves all four adapted RGB+MV detectors by 0.87--2.39 mAP. MiniROAD gives the largest gain and uses only 13.67M auxiliary temporal parameters, versus 88.38M for CMeRT and 56.78M for LSTR/TeSTra (Appendix D.5). This motivates testing whether one lightweight MV-only MiniROAD can serve different joint architectures. Reusing the same trained auxiliary further improves CMeRT, LSTR, and TeSTra by 0.73, 2.15, and 1.51 points over matched routing. The unified system gains 2.22--3.02 points over adapted RGB+MV, approaching or surpassing RGB+Flow accuracy without architecture-specific auxiliary training.

\subsection{Processing Efficiency}
\label{sec:exp_efficiency}

As shown in Table~\ref{tab:efficiency}, CodecMotion-OAD achieves 1.99\(\times\)/11.13\(\times\) GPU/CPU speedups for MiniROAD and 1.58\(\times\)/7.71\(\times\) for CMeRT over TV-L1 RGB+Flow. The unified auxiliary adds only 1.2 GPU ms/step for MiniROAD and 0.8 for CMeRT because motion encoding is shared. TV-L1 dominates CPU preprocessing (\(\sim\)1011 ms/step), versus \(\sim\)6.2--6.3 ms for MV processing. We evaluate against TV-L1; hardware-accelerated Flow such as NVOFA \cite{zhao2022testra} is a separate alternative.



\subsection{Component Analysis}
\label{sec:exp_ablation}

\subsubsection{Flow-to-MV Adaptation}
\label{sec:exp_adaptation}

Table~\ref{tab:adapt} isolates the effect of representation adaptation under the same CMeRT supervised training and Flow-pretrained initialization.

Adaptation raises feature similarity and improves OAD by 4.79 mAP. Appendix A.7 provides additional representation diagnostics.

\subsubsection{Fusion Strategies}
\label{sec:exp_fusion}

Table~\ref{tab:fusion} compares our log-odds presence routing with full-probability averaging and arithmetic presence fusion on the same seven fixed prediction pairs. Both presence-only methods preserve the joint detector's conditional Identity distribution.

Presence-only fusion exceeds probability averaging by 1.23--3.40 points; log-odds adds 0.30--0.94 over arithmetic fusion. Thus, retaining joint Identity predictions is important, and our chosen fusion performs best across the seven pairs. Detailed Presence/Identity results appear in Appendix A.10. Flow-only routing also improves RGB+Flow (Appendix A.9).

\subsubsection{Auxiliary Modality and Complementarity}
\label{sec:exp_aux}

To test whether routing gains merely reflect ensembling, Table~\ref{tab:aux} compares different auxiliary modalities with a fixed MiniROAD RGB+MV detector (seed 20).

\begin{table}[!tbp]
  \centering
  \caption{Fixed-main auxiliary controls (\(\lambda=0.5\)).}
  \label{tab:aux}
  \small
  \setlength{\tabcolsep}{5pt}
  \begin{tabular}{@{}lrrr@{}}
    \hline
    \textbf{Auxiliary} &
    \shortstack{\textbf{Presence}\\\textbf{AP}} &
    \shortstack{\textbf{Probability}\\\textbf{avg.}} &
    \shortstack{\textbf{Routing}\\\textbf{mAP}} \\
    \hline
    RGB+MV (seed 0) & 82.52 & 68.39 & 68.14 \\
    RGB-only        & 76.33 & 66.27 & 65.75 \\
    MV-only         & 79.56 & 66.59 & 69.60 \\
    \hline
  \end{tabular}
\end{table}

Despite lower standalone Presence AP than RGB+MV, MV-only provides the greatest routing gain, indicating useful complementarity beyond a second-model effect. Main/MV-only presence false-positive sets overlap by only 30.2\% Jaccard; routing reduces their false-positive rates to 4.37\% (Appendix A.4). Seed and cross-architecture controls appear in Appendix A.10; \(\lambda\) sensitivity appears in Appendix A.6.

\subsection{Originally Coded Videos}
\label{sec:exp_original}

We also test original videos containing B pictures, retaining only past-directed MVs to reduce future-reference dependence. MV representations and MiniROAD detectors are prepared anew for this coding condition.

\begin{table}[!tbp]
  \centering
  \caption{MiniROAD detection mAP on original coded videos.}
  \label{tab:original}
  \small
  \setlength{\tabcolsep}{20pt}
  \begin{tabular}{@{}lr@{}}
    \hline
    \textbf{Configuration} & \textbf{mAP} \\
    \hline
    RGB+Flow & 70.90 \\
    MV-only & 49.61 \\
    RGB+MV (Adapted) & 67.44 \\
    Probability averaging & 66.38 \\
    CodecMotion-OAD & 69.27 \\
    \hline
  \end{tabular}
\end{table}

As reported in Table~\ref{tab:original}, routing raises original-coded RGB+MV from 67.44 to 69.27 mAP, whereas probability averaging falls to 66.38. This verifies effectiveness beyond controlled IPPP inputs; decoding-reference availability is analyzed separately in Appendix C, with further original-coding diagnostics in Appendix A.8.




\section{Conclusion}
\label{sec:conclusion}

Controlled Flow coarsening shows that fine-grained motion details contribute limited additional accuracy after RGB fusion. Meanwhile, RGB-only detection performs substantially worse, indicating that motion cues remain necessary to maintain high detection accuracy. Therefore, we retain motion cues through inexpensive codec motion vectors (MVs). Visual comparisons show that Flow and MVs preserve similar dominant motion structures, motivating the reuse of Flow-pretrained representations. Our presence--identity analysis shows that RGB substantially narrows the Flow--MV gap in conditional category discrimination, while the action-presence gap remains and motion-only predictions provide complementary presence evidence. Guided by these findings, we propose CodecMotion-OAD, combining unlabeled Flow-to-MV representation adaptation with presence--identity routing to preserve detection accuracy when replacing computationally expensive dense Flow. Experiments across four OAD architectures demonstrate accuracy close to RGB+Flow while substantially reducing complete decoder-to-prediction processing time. We also provide codec MVs and extracted MV features for THUMOS'14 under both original coding and IPPP re-encoding, and analyze how prediction structures affect motion availability in causal online detection.

\clearpage
{
    \small
    \bibliographystyle{ieeenat_fullname}
    \bibliography{main}
}
\clearpage
\clearpage
\setcounter{page}{1}
\hypersetup{pageanchor=false}
\maketitlesupplementary

\appendix
\renewcommand{\theHtable}{supp.\arabic{table}}
\renewcommand{\theHfigure}{supp.\arabic{figure}}
\newsavebox{\suppTableBox}

\section{Additional Experimental Analysis}\label{sup:a}

\subsection{Standalone Motion Discrimination and RGB Fusion}\label{sup:a.1}

\par\addvspace{4pt}\noindent\begin{minipage}{\columnwidth}
\centering

\renewcommand{\thetable}{A1(a)}
\captionof{table}{Detection mAP, Presence AP, and conditional Identity macro-AP for the five MiniROAD inputs.}\label{sup:tab:a1a}

\small\setlength{\tabcolsep}{3pt}
\begin{lrbox}{\suppTableBox}\begin{tabular}{@{}lrrr@{}}\toprule
Input & mAP & Presence AP & \shortstack{Identity macro-AP} \\
\midrule
RGB-only & 59.96 & 76.33 & 94.40 \\
MV-only & 52.52 & 79.56 & 75.97 \\
Flow-only & 67.56 & 83.18 & 91.14 \\
RGB+MV & 67.21 & 79.69 & 95.52 \\
RGB+Flow & 70.39 & 82.79 & 96.52 \\
\bottomrule\end{tabular}\end{lrbox}
\ifdim\wd\suppTableBox>\linewidth
  \resizebox{\linewidth}{!}{\usebox{\suppTableBox}}%
\else
  \usebox{\suppTableBox}%
\fi
\end{minipage}\par\addvspace{4pt}

Official-configuration MiniROAD modality diagnosis on the 211 THUMOS'14 test videos. All models use ground-truth supervision.

\par\addvspace{4pt}\noindent\begin{minipage}{\columnwidth}
\centering

\renewcommand{\thetable}{A1(b)}
\captionof{table}{Presence AUROC and conditional Identity top-1 for the four inputs that include motion.}\label{sup:tab:a1b}

\small\setlength{\tabcolsep}{3pt}
\begin{lrbox}{\suppTableBox}\begin{tabular}{@{}lrr@{}}\toprule
Input & \shortstack{Presence AUROC} & \shortstack{Identity top-1} \\
\midrule
MV-only & 88.07 & 76.90 \\
Flow-only & 90.58 & 86.45 \\
RGB+MV baseline & 88.73 & 89.89 \\
RGB+Flow & 90.42 & 93.01 \\
\bottomrule\end{tabular}\end{lrbox}
\ifdim\wd\suppTableBox>\linewidth
  \resizebox{\linewidth}{!}{\usebox{\suppTableBox}}%
\else
  \usebox{\suppTableBox}%
\fi
\end{minipage}\par\addvspace{4pt}

Differences are computed before rounding and can differ in the last digit from subtraction of displayed values.

\subsection{Sensitivity to the Dense-Flow Estimator}\label{sup:a.2}

Table~\ref{sup:tab:a2} compares TV-L1, RAFT-Large {[}S1{]}, and codec MV with streaming CMeRT. RAFT uses the fixed Flow encoder; MV uses adaptation. Complete processing latency appears in Section 5.3 and Appendix D.

\par\addvspace{4pt}\noindent\begin{minipage}{\columnwidth}
\centering

\renewcommand{\thetable}{A2}
\captionof{table}{CMeRT detection accuracy with three motion sources under streaming evaluation on 211 test videos.}\label{sup:tab:a2}

\small\setlength{\tabcolsep}{3pt}
\begin{lrbox}{\suppTableBox}\begin{tabular}{@{}lrr@{}}\toprule
\shortstack{Motion source} & \shortstack{Feature encoder} & mAP \\
\midrule
TV-L1 & Pretrained Flow-ResNet50 & \textbf{70.50} \\
RAFT-Large & Fixed Flow-ResNet50 & 63.99 \\
Codec MV & Adapted MV encoder & 68.31 \\
\bottomrule\end{tabular}\end{lrbox}
\ifdim\wd\suppTableBox>\linewidth
  \resizebox{\linewidth}{!}{\usebox{\suppTableBox}}%
\else
  \usebox{\suppTableBox}%
\fi
\end{minipage}\par\addvspace{4pt}

With the fixed Flow encoder, RAFT achieves 63.99 mAP; adapting it lowers mAP to 59.91. MV adaptation instead raises mAP from 63.52 to 68.31 under matched detector supervision.

Raw cosine measures direct alignment; centered cosine removes training-set means; ridge regression measures student-to-teacher predictability. All fitting uses training data and is evaluated on held-out data (Table~\ref{sup:tab:a6}).

Table~\ref{sup:tab:a3} reports frame-wise linear-probe results on frozen features (30 epochs, AdamW; averaged over random initializations 0--2).

\par\addvspace{4pt}\noindent\begin{minipage}{\columnwidth}
\centering

\renewcommand{\thetable}{A3}
\captionof{table}{Linear-probe mAP before and after adaptation. TV-L1-only, RGB-only, and RGB+TV-L1 references obtain 31.89, 38.84, and 46.03 mAP, respectively.}\label{sup:tab:a3}

\small\setlength{\tabcolsep}{3pt}
\begin{lrbox}{\suppTableBox}\begin{tabular}{@{}lrrr@{}}\toprule
\shortstack{Feature input} & Fixed mAP & Adapted mAP & Change \\
\midrule
RAFT-Large only & 17.02 & 13.35 & \ensuremath{-}3.67 \\
RGB + RAFT-Large & 41.01 & 40.28 & \ensuremath{-}0.73 \\
Codec MV only & 10.39 & 23.12 & +12.73 \\
RGB + Codec MV & 39.92 & 43.52 & +3.60 \\
\bottomrule\end{tabular}\end{lrbox}
\ifdim\wd\suppTableBox>\linewidth
  \resizebox{\linewidth}{!}{\usebox{\suppTableBox}}%
\else
  \usebox{\suppTableBox}%
\fi
\end{minipage}\par\addvspace{4pt}

\subsection{Feature Predictability and Conditional Utility}\label{sup:a.3}

Probes and regressions use frozen features: 167,037 training feature steps (164,833 non-ambiguous) and 183,266 test steps (180,400 non-ambiguous). Standardization is fitted on training data.

The frame-wise probe predicts 20 actions plus background using class-weighted cross-entropy (background weight 0.5) and AdamW for 30 epochs (\(10^{-3}\) learning rate, \(10^{-2}\) weight decay, batch 4096); mAP is averaged over three seeds. No temporal context is used.

On standardized features, ridge regression uses coefficient \(0.01N\) for \(N\) training frames and pooled test-set \(R^2\). A train-fitted MV-to-Flow map defines the Flow residual; probes evaluate that residual as an extra feature.

Probes confirm that RGB contracts the Flow--MV mAP gap from 8.8 to 2.5 points, independent of temporal modeling.

Feature overlap remains limited: MV regression leaves about 59\% of Flow variance unexplained, and RGB predicts that residual poorly (\(R^2=0.026\)).

The Flow residual improves MV-only probe mAP from 23.1 to 31.5 but RGB+MV from 43.5 to just 45.4, supporting its lower conditional task utility after RGB fusion.

\subsection{Per-Class, Presence/Identity, and Boundary Analyses}\label{sup:a.4}

RGB reduces the Flow--MV AP gap in 18 of 20 classes and reverses it in four. Mean per-class gaps fall from 15.04 to 3.19 points; the latter reflects unrounded diagnostic values. Largest contractions are tabulated below.

\par\addvspace{4pt}\noindent\begin{minipage}{\columnwidth}
\centering

{\small Largest per-class contractions of the Flow-\/-MV gap after RGB fusion.\par}\smallskip

\small\setlength{\tabcolsep}{3pt}
\begin{lrbox}{\suppTableBox}\begin{tabular}{@{}lrrr@{}}\toprule
Class & \shortstack{Motion-only gap} & +RGB gap & Contraction \\
\midrule
HighJump & +37.3 & +8.0 & +29.4 \\
BaseballPitch & +14.1 & \ensuremath{-}9.3 & +23.4 \\
FrisbeeCatch & +28.0 & +4.7 & +23.3 \\
LongJump & +18.9 & \ensuremath{-}2.6 & +21.5 \\
SoccerPenalty & +22.7 & +3.6 & +19.1 \\
\bottomrule\end{tabular}\end{lrbox}
\ifdim\wd\suppTableBox>\linewidth
  \resizebox{\linewidth}{!}{\usebox{\suppTableBox}}%
\else
  \usebox{\suppTableBox}%
\fi
\end{minipage}\par\addvspace{4pt}

HammerThrow and VolleyballSpiking have contractions of \ensuremath{-}0.2 and \ensuremath{-}1.5 points. The motion-only and post-fusion gaps are moderately correlated across classes (Spearman \(\rho=0.56\), \(p=0.01\)).

The Flow--MV gaps in detection, presence, and identity appear in Table 2. RGB fusion reduces the identity macro-AP gap by approximately 93\%, compared with about 14\% for presence AP. The corresponding AUROC and top-1 results follow the same pattern.

We group frames by distance to the nearest action/background transition on the 4-fps feature grid. Replacing only the joint MV predictions within 0--7 steps of a transition with joint flow predictions raises global detection mAP by 1.89 points; this region covers approximately 36\% of frames. The replacement uses predictions already available from the two models and is a diagnostic intervention, not an inference rule.

\begin{table*}[!htbp]
\centering

{\small Presence and prediction replacement by distance to an action/background transition.\par}\smallskip

\small\setlength{\tabcolsep}{3pt}
\begin{lrbox}{\suppTableBox}\begin{tabular}{@{}lrrrrrr@{}}\toprule
\shortstack{Distance (steps)} & \shortstack{Frame share (\%)} & \shortstack{Action share\\within group (\%)} & \shortstack{MV Presence AP} & \shortstack{Flow Presence AP} & Difference & \shortstack{Replacement mAP\\gain} \\
\midrule
0--1 & 11.8 & 52.2 & 67.98 & 69.53 & +1.55 & +0.43 \\
2--3 & 9.9 & 54.2 & 84.86 & 86.79 & +1.93 & +0.79 \\
4--7 & 14.4 & 51.8 & 89.34 & 91.08 & +1.74 & +0.67 \\
8--15 & 17.8 & 45.9 & 91.64 & 93.37 & +1.73 & \ensuremath{-}0.19 \\
\ensuremath{\geq}16 & 46.0 & 11.4 & 65.02 & 70.77 & +5.75 & \ensuremath{-}0.93 \\
\bottomrule\end{tabular}\end{lrbox}
\ifdim\wd\suppTableBox>\linewidth
  \resizebox{\linewidth}{!}{\usebox{\suppTableBox}}%
\else
  \usebox{\suppTableBox}%
\fi
\end{table*}

Replacement gains interact through global AP ranking and are nonadditive. The combined 0--7-step intervention yields +1.89 mAP, whereas replacing distant background-dominated frames alone reduces mAP. This localizes where flow predictions are most useful in this diagnostic.

For the fixed main and auxiliary pair, presence errors overlap as follows. Miss rates use action frames as denominator; false-positive rates use background frames. Conditional overlap uses the main model\textquotesingle s error set as denominator, while Jaccard overlap uses the union. The F1-optimal thresholds are selected separately on these diagnostic predictions (main 0.128, auxiliary 0.127, routing 0.158).

\begin{table*}[!htbp]
\centering

{\small Overlap of presence errors for the MiniROAD prediction pair.\par}\smallskip

\small\setlength{\tabcolsep}{3pt}
\begin{lrbox}{\suppTableBox}\begin{tabular}{@{}lrr@{}}\toprule
Statistic & \shortstack{Threshold 0.5} & \shortstack{Separately\\F1-optimal\\thresholds} \\
\midrule
Miss rate: main / auxiliary / routing (\%) & 44.47 / 45.42 / 45.81 & 22.57 / 22.50 / 23.12 \\
False-positive rate: main / auxiliary / routing (\%) & 5.60 / 6.15 / 4.37 & 17.07 / 18.09 / 12.89 \\
Auxiliary also misses given main miss (\%) & 73.9 & 57.5 \\
Miss-set Jaccard (\%) & 57.6 & 40.4 \\
Auxiliary also errs given main false positive (\%) & 48.7 & 54.4 \\
False-positive-set Jaccard (\%) & 30.2 & 35.9 \\
\bottomrule\end{tabular}\end{lrbox}
\ifdim\wd\suppTableBox>\linewidth
  \resizebox{\linewidth}{!}{\usebox{\suppTableBox}}%
\else
  \usebox{\suppTableBox}%
\fi
\end{table*}

False positives overlap less than misses under both threshold conventions. Routing lowers false-positive rates while slightly increasing misses. Feature smoothness diagnostics separately give adjacent-frame cosine values of 0.913 for MV, 0.837 for RGB, and 0.778 for Flow.

\subsection{Controlled Flow Spatial-Coarsening Intervention}\label{sup:a.5}

We coarsen native \(320\times180\) TV-L1 Flow by replacing each \(g\times g\) block (\(g\in\{8,16,32\}\)) with its mean 2-D vector, tiling back before standard preprocessing. Partial boundary blocks use valid pixels; a 1\ensuremath{\times}1 control exactly reproduces original Flow (cosine 1.0000).

For each block size, frozen and adapted Flow-ResNet50 encoders are tested with motion-only and RGB+motion MiniROAD: 12 models, and 211 test videos. Original Flow and codec MV rows reuse their matched runs.

Table~\ref{sup:tab:a4a}--\ref{sup:tab:a4b} reports detection and factor-level metrics for motion-only and RGB+motion inputs. Feature similarities for the adapted encoders are given in Table 1.

\par\addvspace{4pt}\noindent\begin{minipage}{\columnwidth}
\centering

\renewcommand{\thetable}{A4(a)}
\captionof{table}{Official MiniROAD results under spatial coarsening with frozen and adapted motion encoders. The RGB+motion Codec MV row is RGB+MV baseline. Presence and identity metrics follow Section 3.3. Motion-only inputs.}\label{sup:tab:a4a}

\small\setlength{\tabcolsep}{3pt}
\begin{lrbox}{\suppTableBox}\begin{tabular}{@{}lrrrrr@{}}\toprule
Input / encoder & mAP & \shortstack{Presence\\AP} & \shortstack{Presence\\AUROC} & \shortstack{Identity\\macro-AP} & \shortstack{Identity\\top-1} \\
\midrule
Original Flow & 67.56 & 83.18 & 90.58 & 91.14 & 86.45 \\
8\ensuremath{\times}8 / frozen & 30.47 & 69.14 & 80.84 & 48.82 & 52.67 \\
16\ensuremath{\times}16 / frozen & 27.58 & 64.97 & 79.10 & 46.60 & 48.19 \\
32\ensuremath{\times}32 / frozen & 26.12 & 60.62 & 74.23 & 45.43 & 51.45 \\
8\ensuremath{\times}8 / adapted & 65.60 & 84.15 & 90.40 & 89.80 & 88.32 \\
16\ensuremath{\times}16 / adapted & 57.27 & 81.17 & 88.32 & 81.14 & 82.72 \\
32\ensuremath{\times}32 / adapted & 42.26 & 75.86 & 84.55 & 64.55 & 65.66 \\
Codec MV / adapted & 52.52 & 79.56 & 88.07 & 75.97 & 76.90 \\
\bottomrule\end{tabular}\end{lrbox}
\ifdim\wd\suppTableBox>\linewidth
  \resizebox{\linewidth}{!}{\usebox{\suppTableBox}}%
\else
  \usebox{\suppTableBox}%
\fi
\end{minipage}\par\addvspace{4pt}

\par\addvspace{4pt}\noindent\begin{minipage}{\columnwidth}
\centering

\renewcommand{\thetable}{A4(b)}
\captionof{table}{Official MiniROAD results under spatial coarsening with frozen and adapted motion encoders. The RGB+motion Codec MV row is RGB+MV baseline. Presence and identity metrics follow Section 3.3. RGB+motion inputs.}\label{sup:tab:a4b}

\small\setlength{\tabcolsep}{3pt}
\begin{lrbox}{\suppTableBox}\begin{tabular}{@{}lrrrrr@{}}\toprule
Input / encoder & mAP & \shortstack{Presence\\AP} & \shortstack{Presence\\AUROC} & \shortstack{Identity\\macro-AP} & \shortstack{Identity\\top-1} \\
\midrule
Original Flow & 70.39 & 82.79 & 90.42 & 96.52 & 93.01 \\
8\ensuremath{\times}8 / frozen & 63.29 & 78.51 & 87.96 & 93.91 & 90.46 \\
16\ensuremath{\times}16 / frozen & 62.96 & 78.93 & 88.02 & 94.15 & 90.94 \\
32\ensuremath{\times}32 / frozen & 62.60 & 78.62 & 88.13 & 94.62 & 90.95 \\
8\ensuremath{\times}8 / adapted & 70.19 & 83.04 & 90.36 & 95.83 & 90.88 \\
16\ensuremath{\times}16 / adapted & 68.83 & 80.73 & 89.76 & 95.67 & 91.62 \\
32\ensuremath{\times}32 / adapted & 65.43 & 79.05 & 87.82 & 94.67 & 89.97 \\
Codec MV / adapted & 67.21 & 79.69 & 88.73 & 95.52 & 89.89 \\
\bottomrule\end{tabular}\end{lrbox}
\ifdim\wd\suppTableBox>\linewidth
  \resizebox{\linewidth}{!}{\usebox{\suppTableBox}}%
\else
  \usebox{\suppTableBox}%
\fi
\end{minipage}\par\addvspace{4pt}

Feature cosine is evaluated on the 211 test videos using the same procedure for all adapted encoders. The codec-motion value of 0.828 corresponds to the adapted MV representation in main-text Table 5. Encoder training uses the training split only.

Adapting Flow-initialized encoders substantially improves all coarsened inputs (Table~\ref{sup:tab:a4a}--\ref{sup:tab:a4b}): motion-only gains are 35.13, 29.69, and 16.14 mAP at 8\ensuremath{\times}8, 16\ensuremath{\times}16, and 32\ensuremath{\times}32; RGB-fused gains are 6.90, 5.87, and 2.83. Encoder adaptation is therefore important even when coarse motion retains useful structure.

Linear probes show the same interaction: coarsening reduces motion-only identity more than presence, while RGB greatly reduces the identity loss. The codec MV probe follows this trend.

Controlled coarsening shows that fine Flow detail contributes relatively little additional two-stream accuracy, while adaptation makes coarser motion usable. Together with Figure 2, these results motivate Flow-pretrained MV initialization followed by representation adaptation; the actual MV effect is measured in main-text Table 5.

\subsection{Presence-Weight Sensitivity}\label{sup:a.6}

We evaluate five auxiliary-presence weights with a fixed MiniROAD RGB+MV main detector and MV-only auxiliary. Equal weighting (\(\lambda=0.5\)) yields the highest mAP and Presence AP in this sweep. With \(\lambda=0\), output reconstruction still sets the ambiguous channel to zero, giving 67.23 mAP rather than the direct baseline output of 67.21 mAP.

\par\addvspace{4pt}\noindent\begin{minipage}{\columnwidth}
\centering

\renewcommand{\thetable}{A5}
\captionof{table}{Presence--Identity Routing weight sensitivity for the fixed MiniROAD prediction pair. Identity macro-AP is unchanged across weights.}\label{sup:tab:a5}

\small\setlength{\tabcolsep}{3pt}
\begin{lrbox}{\suppTableBox}\begin{tabular}{@{}lrrrrr@{}}\toprule
\shortstack{Auxiliary weight\\\(\lambda\)} & 0 & 0.25 & 0.5 & 0.75 & 1 \\
\midrule
mAP & 67.23 & 69.41 & \textbf{69.60} & 67.67 & 64.11 \\
Presence AP & 79.69 & 81.74 & \textbf{82.52} & 81.76 & 79.56 \\
Presence AUROC & 88.73 & 89.98 & \textbf{90.38} & 89.71 & 88.07 \\
Identity macro-AP & 95.52 & 95.52 & 95.52 & 95.52 & 95.52 \\
\bottomrule\end{tabular}\end{lrbox}
\ifdim\wd\suppTableBox>\linewidth
  \resizebox{\linewidth}{!}{\usebox{\suppTableBox}}%
\else
  \usebox{\suppTableBox}%
\fi
\end{minipage}\par\addvspace{4pt}

\subsection{Representation Adaptation Diagnostics Across Motion Sources}\label{sup:a.7}

We evaluate MV and RAFT-Large {[}S1{]} adaptation using the same TV-L1-pretrained initialization, frozen teacher, five-field inputs, and supervised CMeRT training. Raw cosine is direct alignment, centered cosine subtracts the training-set feature means, and \(R^2\) is the held-out ridge-regression score.

\par\addvspace{4pt}\noindent\begin{minipage}{\columnwidth}
\centering

\renewcommand{\thetable}{A6}
\captionof{table}{Feature alignment, feature predictability, and RGB+motion CMeRT detection mAP for fixed and adapted RAFT/MV encoders (211-video streaming evaluation).}\label{sup:tab:a6}

\small\setlength{\tabcolsep}{3pt}
\begin{lrbox}{\suppTableBox}\begin{tabular}{@{}lrrrr@{}}\toprule
Source/state & Raw cos. & Ctr. cos. & \(R^2\) & mAP \\
\midrule
RAFT fixed & 0.578 & 0.306 & 0.290 & 63.99 \\
RAFT adapted & 0.721 & 0.277 & 0.232 & 59.91 \\
MV fixed & 0.456 & 0.066 & 0.206 & 63.52 \\
MV adapted & \textbf{0.828} & \textbf{0.623} & \textbf{0.407} & \textbf{68.31} \\
\bottomrule\end{tabular}\end{lrbox}
\ifdim\wd\suppTableBox>\linewidth
  \resizebox{\linewidth}{!}{\usebox{\suppTableBox}}%
\else
  \usebox{\suppTableBox}%
\fi
\end{minipage}\par\addvspace{4pt}

RAFT adaptation improves raw cosine but reduces centered cosine, \(R^2\), and detection mAP (63.99 to 59.91), whereas MV adaptation improves each metric (63.52 to 68.31 mAP). The comparison characterizes source-dependent representation adaptation.

\subsection{Additional Original-Coding Results}\label{sup:a.8}

The original-coded condition extracts past-directed vectors from B pictures; IPPP uses past-referenced coding. Main-text Table 8 reports MiniROAD results on original coded videos. Table~\ref{sup:tab:a7} pairs the original-coding measurements with IPPP results, and Table~\ref{sup:tab:a8} gives the RGB+Flow references for all four detectors. Encoder cosine is measured on validation features, while detection mAP uses 211 test videos.

\par\addvspace{4pt}\noindent\begin{minipage}{\columnwidth}
\centering

\renewcommand{\thetable}{A7}
\captionof{table}{MV feature and prediction diagnostics under original coding versus IPPP.}\label{sup:tab:a7}

\small\setlength{\tabcolsep}{3pt}
\begin{lrbox}{\suppTableBox}\begin{tabular}{@{}lrr@{}}\toprule
Metric & \shortstack{Original coded\\video} & IPPP \\
\midrule
Encoder validation cosine & 0.800 & 0.815 \\
MV-only mAP & 49.61 & 52.52 \\
RGB+MV baseline mAP & 67.44 & 67.21 \\
Probability averaging mAP & 66.38 & --- \\
Presence--identity routing mAP & \textbf{69.27} & \textbf{69.60} \\
\bottomrule\end{tabular}\end{lrbox}
\ifdim\wd\suppTableBox>\linewidth
  \resizebox{\linewidth}{!}{\usebox{\suppTableBox}}%
\else
  \usebox{\suppTableBox}%
\fi
\end{minipage}\par\addvspace{4pt}

\par\addvspace{4pt}\noindent\begin{minipage}{\columnwidth}
\centering

\renewcommand{\thetable}{A8}
\captionof{table}{RGB+Flow references on original coded and IPPP THUMOS\textquotesingle14 videos for four temporal detectors (211 test videos); CMeRT uses streaming evaluation.}\label{sup:tab:a8}

\small\setlength{\tabcolsep}{3pt}
\begin{lrbox}{\suppTableBox}\begin{tabular}{@{}lrr@{}}\toprule
Detector & \shortstack{Original coded\\video} & \shortstack{IPPP reference} \\
\midrule
MiniROAD & 70.90 & 70.39 \\
CMeRT & 70.95 & 70.50 \\
LSTR & 70.16 & 69.60 \\
TeSTra & 70.40 & 70.50 \\
\bottomrule\end{tabular}\end{lrbox}
\ifdim\wd\suppTableBox>\linewidth
  \resizebox{\linewidth}{!}{\usebox{\suppTableBox}}%
\else
  \usebox{\suppTableBox}%
\fi
\end{minipage}\par\addvspace{4pt}

\subsection{Presence--Identity Routing with Optical Flow}\label{sup:a.9}

We apply the same routing method to a fixed MiniROAD RGB+Flow main detector and Flow-only auxiliary, with prediction pairs fixed within each coding setting.

\par\smallskip\noindent\begin{minipage}{\columnwidth}
\centering

\renewcommand{\thetable}{A9}
\captionof{table}{Flow-only routing on original coded and IPPP THUMOS\textquotesingle14 videos (211 test videos); IPPP routing includes three auxiliary-presence weights.}\label{sup:tab:a9}

\small\setlength{\tabcolsep}{3pt}
\begin{lrbox}{\suppTableBox}\begin{tabular}{@{}lrr@{}}\toprule
Configuration & Original coding & IPPP \\
\midrule
RGB+Flow main & 70.90 & 70.39 \\
Flow-only auxiliary & 65.83 & 67.56 \\
Probability averaging & 72.18 & 72.53 \\
Routing (original coding) & \textbf{73.22} & --- \\
Routing ($\lambda=0.25$) & --- & 72.12 \\
Routing ($\lambda=0.5$) & --- & \textbf{72.76} \\
Routing ($\lambda=0.75$) & --- & 72.09 \\
\bottomrule\end{tabular}\end{lrbox}
\ifdim\wd\suppTableBox>\linewidth
  \resizebox{\linewidth}{!}{\usebox{\suppTableBox}}%
\else
  \usebox{\suppTableBox}%
\fi
\end{minipage}\par\smallskip

On original coded videos, routing achieves 73.22 mAP versus 70.90 for the main detector and 72.18 for averaging. On IPPP, equal-weight routing achieves 72.76 versus 70.39 for the main detector and 72.53 for averaging. All three tested IPPP weights improve over the main detector, with the largest gain at \(\lambda=0.5\).

\subsection{Additional Routing Fusion and Auxiliary Controls}\label{sup:a.10}

\textbf{Presence-only fusion versus full-probability averaging.} We compare the routing method with arithmetic presence interpolation, \(a_t^{\mathrm{arith}}=(1-\lambda)a_t^j+\lambda a_t^m\), followed by the same conditional category distribution \(q_{t,c}\) of the joint detector. main-text Table 6 reports the seven detection mAP comparisons. Table~\ref{sup:tab:a10} preserves the Presence AP and conditional Identity macro-AP measurements from the same fixed prediction pairs.

\par\smallskip\noindent\begin{minipage}{\columnwidth}
\centering

\renewcommand{\thetable}{A10}
\captionof{table}{Detection mAP, Presence AP and Identity macro-AP for seven main / MV-only pairs ($\lambda=0.5$, 211 test videos). Avg. denotes probability averaging.}\label{sup:tab:a10}

\small\setlength{\tabcolsep}{3pt}
\begin{lrbox}{\suppTableBox}\begin{tabular}{@{}llrrr@{}}\toprule
Detector pair & Measure & Baseline & Avg. & Routing \\
\midrule
MiniROAD / & mAP & 67.21 & 66.59 & \textbf{69.60} \\
MiniROAD & Pres. AP & 79.69 & 82.04 & \textbf{82.52} \\
 & Id. mAP & 95.52 & 92.91 & \textbf{95.52} \\
\addlinespace[1pt]
CMeRT / & mAP & 68.31 & 67.99 & \textbf{69.80} \\
CMeRT & Pres. AP & 81.88 & 83.81 & \textbf{84.02} \\
 & Id. mAP & 95.57 & 94.35 & \textbf{95.57} \\
\addlinespace[1pt]
CMeRT / & mAP & 68.31 & 68.91 & \textbf{70.53} \\
MiniROAD & Pres. AP & 81.88 & 83.71 & \textbf{84.06} \\
 & Id. mAP & 95.57 & 94.60 & \textbf{95.57} \\
\addlinespace[1pt]
LSTR / & mAP & 66.74 & 63.46 & \textbf{67.61} \\
LSTR & Pres. AP & 78.94 & 81.14 & \textbf{81.35} \\
 & Id. mAP & 95.18 & 91.92 & \textbf{95.18} \\
\addlinespace[1pt]
LSTR / & mAP & 66.74 & 67.47 & \textbf{69.76} \\
MiniROAD & Pres. AP & 78.94 & 82.33 & \textbf{82.58} \\
 & Id. mAP & 95.18 & 93.69 & \textbf{95.18} \\
\addlinespace[1pt]
TeSTra / & mAP & 68.20 & 66.01 & \textbf{69.60} \\
TeSTra & Pres. AP & 82.80 & 83.89 & \textbf{84.33} \\
 & Id. mAP & 95.73 & 92.19 & \textbf{95.73} \\
\addlinespace[1pt]
TeSTra / & mAP & 68.20 & 68.03 & \textbf{71.11} \\
MiniROAD & Pres. AP & 82.80 & 83.64 & \textbf{84.47} \\
 & Id. mAP & 95.73 & 93.46 & \textbf{95.73} \\
\addlinespace[1pt]
\bottomrule\end{tabular}\end{lrbox}
\ifdim\wd\suppTableBox>\linewidth
  \resizebox{\linewidth}{!}{\usebox{\suppTableBox}}%
\else
  \usebox{\suppTableBox}%
\fi
\end{minipage}\par\smallskip

\textbf{Auxiliary-input controls across detectors.} Table~\ref{sup:tab:a11} extends the MiniROAD controls of main-text Table 7 to four main detector architectures. For the second RGB+MV and RGB-only auxiliaries, we report the higher mAP of probability averaging and presence routing at \(\lambda=0.5\); the MV-only column uses the complete CodecMotion-OAD routing configuration.

\par\addvspace{4pt}\noindent\begin{minipage}{\columnwidth}
\centering

\renewcommand{\thetable}{A11}
\captionof{table}{Comparison of MV-only routing with the best evaluated fusion of a second RGB+MV or RGB-only auxiliary; all values are detection mAP (211 test videos).}\label{sup:tab:a11}

\small\setlength{\tabcolsep}{3pt}
\begin{lrbox}{\suppTableBox}\begin{tabular}{@{}lrrr@{}}\toprule
\shortstack{Main detector} & \shortstack{MV-only routing\\(ours)} & \shortstack{Second RGB+MV\\auxiliary (best\\fusion)} & \shortstack{RGB-only\\auxiliary (best\\fusion)} \\
\midrule
MiniROAD & \textbf{69.60} & 68.39 & 66.27 \\
CMeRT & \textbf{70.53} & 69.87 & 66.90 \\
LSTR & \textbf{69.76} & 69.62 & 66.99 \\
TeSTra & \textbf{71.11} & 70.09 & 67.62 \\
\bottomrule\end{tabular}\end{lrbox}
\ifdim\wd\suppTableBox>\linewidth
  \resizebox{\linewidth}{!}{\usebox{\suppTableBox}}%
\else
  \usebox{\suppTableBox}%
\fi
\end{minipage}\par\addvspace{4pt}

MV-only routing gives the highest mAP in all four comparisons. With a second RGB+MV auxiliary, probability averaging can outperform routing.

\textbf{Background-score diagnostic.} All seven pairs retain baseline mAP when only the background score changes.

\twocolumn[{
\begin{minipage}{\textwidth}
\section{Qualitative Analysis of Codec Motion}\label{sup:b}

Figure~\ref{sup:fig:b1} shows aligned TV-L1 and MV fields for body, camera, and localized motion, plus an I picture without MVs. Shared dominant patterns motivate Flow-based initialization; block sparsity motivates adaptation. Visualized MVs are sign-flipped to match Flow direction, with shared per-row magnitude caps (99th percentile of Flow).

\par\medskip\begingroup
\centering
\includegraphics[width=.81\textwidth]{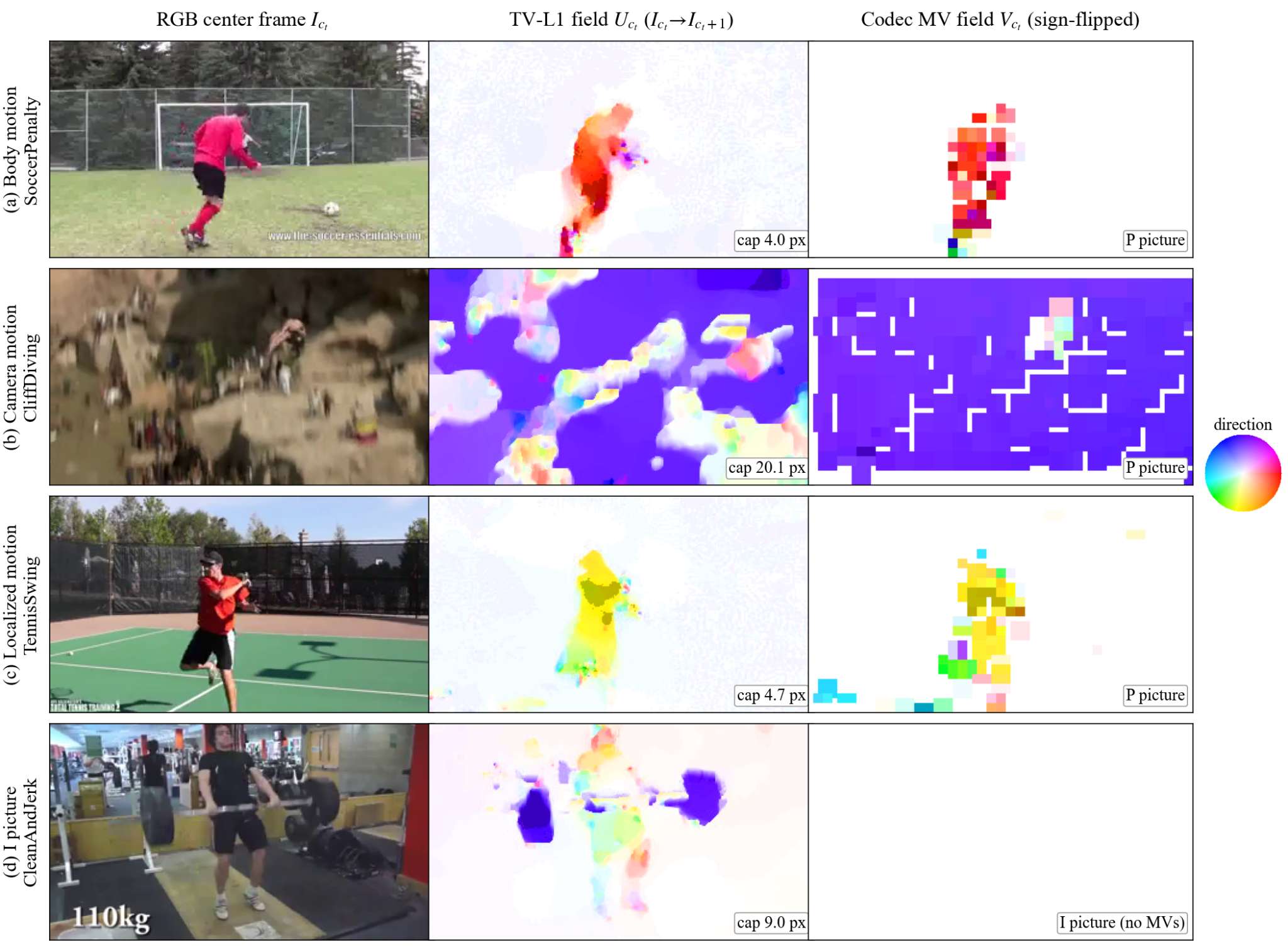}
\renewcommand{\thefigure}{B1}
\captionof{figure}{Qualitative comparison of aligned RGB frames, TV-L1 optical flow, and codec motion vectors from THUMOS'14 under the shared IPPP protocol. The examples show (a) body motion (SoccerPenalty), (b) camera motion (CliffDiving), (c) localized motion (TennisSwing), and (d) an I picture without motion vectors (CleanAndJerk). For visualization, codec MVs are sign-flipped to match the forward-flow direction convention. Within each row, Flow and MV use the same magnitude cap, determined by the 99th percentile of Flow magnitude and labeled in pixels. The displayed motion fields are aligned at \(c_t\) and each constitutes one field of the five-field encoder input.}\label{sup:fig:b1}
\par\endgroup
\end{minipage}\vspace{1em}}]

\section{Coding and Feature Extraction}\label{sup:c}

\subsection{Encoding Choice and Shared IPPP Configuration}\label{sup:c.1}

All controlled sources use H.264/AVC {[}S2{]} re-encoding with \texttt{libx264}, no B pictures, GOP 25, and CRF 23. RGB, Flow, and MVs are extracted from the same IPPP stream.

Past-referenced IPPP limits decoding dependencies to earlier display-order pictures; no picture beyond the selected MV window is required. This aligns with low-delay decoding {[}S3, S4{]}; the original B-picture audit follows.

Among 411 original THUMOS\textquotesingle14 videos (200 train, 211 test), 1,039,610 pictures are B pictures. Their nearest subsequent I/P anchors are 1, 2, or 3 frames ahead in 69.0\%, 19.2\%, and 11.8\% of cases (mean 1.43). An anchor in the following feature interval need not reach the next feature center.

Across 350,303 feature steps, 8.8\% have estimated B-picture dependencies extending beyond the midpoint between feature centers; none reaches the next center. A selected window ending at \(c_t+2\) can require source frame \(c_t+5\). These are decoding dependencies, distinct from detector context and measured processing latency.

Appendix C.3 formalizes source-frame availability; Appendix D defines the associated decoder-to-prediction processing measurement.

\subsection{Motion-Field Extraction and Feature-Grid Alignment}\label{sup:c.2}

PyAV provides libavcodec MV side data. We convert vectors to pixel displacements, rasterize over destination blocks, and zero-fill intra/missing regions. Residuals are not used.

Let a video contain \(N\) source frames \(I_0,\ldots,I_{N-1}\), and let \(T\) be the number of RGB feature steps. We use the existing RGB feature length to define a shared nominal 4-fps grid:

\[(c_0,\ldots,c_{T-1})
=
Q\!\left(\operatorname{linspace}(0,N-1,T)\right),\]

where \(Q\) denotes round-to-nearest followed by integer conversion. RGB, Flow, and MV inputs share these sampling centers. The centers define feature alignment only: the corresponding feature becomes usable after the observations required by its motion extractor are available, as formalized in Appendix C.3. For THUMOS'14, the released \texttt{target\_perframe} has \(T\times22\) columns (background, 20 actions, Ambiguous) and uses nominal quarter-second label intervals \([t/4,(t+1)/4)\). On the 410 videos with available annotation JSON within the 411-video collection, reconstruction agrees with 97.30\% of non-ambiguous labels at zero shift (versus 94.35\% at a \ensuremath{-}1-step shift and 96.64\% at a +1-step shift). Residual discrepancies include overlapping CliffDiving/Diving labels and interval-boundary conventions. The evaluated IPPP RGB/Flow/MV feature lengths match the target axis, and the annotation reconstruction approximates the released label-generation procedure. This full-video grid defines offline benchmark alignment; live deployment would require timestamp-based sampling. Among the 411 training and test videos, 380 are 30 fps, 18 are 29.97 fps, 11 are 25 fps and two are 24 fps; the average selected step spacing exceeds 250 ms by 0.84 ms (range 0.04--8.82 ms), producing cumulative drift relative to regular 4-Hz label intervals. This grid construction is distinct from detector-internal long-memory sampling.

Flow and codec-motion inputs aggregate the same five consecutive fields ending at \(e_t=c_t+2\), indexed by \(e_t-4,\ldots,e_t\). The paired RGB frame is taken two source frames earlier, \(I_{e_t-2}=I_{c_t}\); out-of-range motion indices are clipped to legal endpoints.

MVs are rasterized at native resolution, bilinearly resized to \(180\times320\), then subsampled by two and cached as float16 \([T,5,2,90,160]\). Displacement magnitudes are not geometrically rescaled.

Before the motion encoder, a scale factor of 16 maps displacement values to the flow-JPEG-style intensity convention, e.g.

\[d_x\mapsto\mathrm{clip}(16d_x+128,0,255),\]

followed by spatial interpolation/cropping to the network input size. This is an intensity encoding rather than geometric displacement rescaling.

\subsection{Source-Frame Availability}\label{sup:c.3}

For feature center \(c_t\) and terminal field index \(e_t=c_t+2\), \(\tau_t^X\) is the earliest source-frame index by which modality \(X\) has all observations for that feature step. It excludes subsequent processing time.

For Flow, \(U_i\) is computed from \(I_i\rightarrow I_{i+1}\). With valid Flow indices \(0,\ldots,N_F-1\),

\[\tau_t^F=\max_{j=-4,\ldots,0}\left[\operatorname{clip}(e_t+j,0,N_F-1)+1\right].\]

An interior window therefore requires source frame \(I_{e_t+1}\). Endpoint repetition uses the source observations required by the clipped fields.

For codec motion, define the clipped field set

\[\mathcal{W}_t=\{a_{t,j}\mid j\in\{-4,-3,-2,-1,0\}\},\]

where \(a_{t,j}=\operatorname{clip}(e_t+j,0,N-1)\). Let \(\mathcal{R}^{*}(i)\) be the transitive set of source pictures required to decode picture \(i\), including \(i\) itself. Then

\[\tau_t^M=\max_{i\in\mathcal{W}_t}\max_{r\in\mathcal{R}^{*}(i)}r.\]

Under the past-referenced IPPP configuration, all references precede the coded picture in display order. Since \(\mathcal{R}^{*}(i)\) includes \(i\),

\[\max_{r\in\mathcal{R}^{*}(i)}r=i,
\qquad
\tau_t^M=\max_{i\in\mathcal{W}_t}i\le e_t.\]

Thus, codec references add no source-picture dependency beyond the explicitly selected MV window.

Inference becomes eligible once the required motion observations through \(\tau_t^X\) are available; the resulting motion input is paired with the buffered RGB frame from two source frames earlier, \(I_{e_t-2}=I_{c_t}\). On 350,303 feature steps, terminal-frame availability falls after the nominal label interval at 37.8\% of MV steps (p95 +167 ms) and 49.2\% of TV-L1 steps (p95 +200 ms). Appendix D reports processing latency separately.

\subsection{Training Details}\label{sup:c.4}

MV representation adaptation uses AdamW for 12 epochs (learning rate \(10^{-4}\), weight decay \(10^{-4}\), cosine schedule, batch 24), sampling 32 steps per training video per epoch. BatchNorm is frozen; inputs use fixed resize and 224\ensuremath{\times}224 center crop without augmentation. The final encoder is frozen for ground-truth-supervised main and auxiliary detector training. CMeRT, LSTR, and TeSTra use a 10-epoch linear warm-up (0.3 learning-rate multiplier) followed by cosine decay; detector settings appear below.

\begin{table*}[!htbp]
\centering

\renewcommand{\thetable}{C1}
\caption{Temporal detector training configurations.}\label{sup:tab:c1}

\small\setlength{\tabcolsep}{3pt}
\begin{lrbox}{\suppTableBox}\begin{tabular}{@{}llrrrllp{.25\linewidth}@{}}\toprule
Detector & Optimizer & \shortstack{Learning rate} & Weight decay & Epochs & Batch & Schedule & \shortstack{Configuration} \\
\midrule
MiniROAD & AdamW & 1e-4 & 0.05 & 10 & 16 & None & Window 128; stride 4; embedding 2048; GRU hidden 1024; dropout 0.2 \\
CMeRT & Adam & 2e-4 & 5e-5 & 16 & 32 & Warm-up + cosine & Long memory 256; work memory 4 \\
LSTR & Adam & 7e-5 & 5e-5 & 25 & 16 & Warm-up + cosine & Long memory 512; work memory 8 \\
TeSTra & Adam & 7e-5 & 5e-5 & 25 & 16 & Warm-up + cosine & Long memory 512; work memory 8; decay 0.97 \\
\bottomrule\end{tabular}\end{lrbox}
\ifdim\wd\suppTableBox>\linewidth
  \resizebox{\linewidth}{!}{\usebox{\suppTableBox}}%
\else
  \usebox{\suppTableBox}%
\fi
\end{table*}

MiniROAD uses the official NONUNIFORM objective on THUMOS.

\section{Unified Complete-System Timing}\label{sup:d}

\subsection{Configuration and Reproduction}\label{sup:d.1}

The unified v3 benchmark uses a Tesla V100-SXM2-32GB and an Intel Xeon Gold 6240 @ 2.60 GHz with 16 cores. Software versions are PyTorch 2.1.0+cu118, cuDNN 8.7, PyAV 15.1.0, OpenCV 4.11.0, and NVIDIA driver 550.144.03. GPU runs use four PyTorch CPU threads. CPU runs allocate eight threads to both PyTorch and the resident TV-L1 process. The tested GPU is exclusive to the benchmark; the CPU is otherwise largely idle apart from one external process occupying one core. All models use fp32, batch-one inference.

The benchmark uses the 20 videos listed in latency/sessions20.txt. Each pipeline runs five warm-up and 25 measured steps per video, totaling 500 measured steps. The ten combinations of two detectors and five pipelines rotate their execution order by video index. Each step is timed from the preceding step\textquotesingle s completion until its final class-probability vector is available in host memory. Model loading and resident TV-L1 process startup are excluded. GPU and CPU TV-L1 use resident tvl1\_server and tvl1\_server\_cpu processes, respectively, and compute five flow fields per step.

MiniROAD uses its official RGB-only, RGB+MV baseline, MV-only, and RGB+Flow checkpoints; its update is feature embedding, GRU, ReLU, and classification. CMeRT uses its RGB-only, RGB+MV baseline, and RGB+Flow checkpoints. Both complete systems share the same MiniROAD MV-only auxiliary. The main and auxiliary models share input features, run sequentially on the same device, and return scores to the host for NumPy routing with auxiliary weight 0.5. No parallel CUDA streams are used.

The benchmark is implemented in latency/bench\_unified\_v3.py and launched by latency/run\_all\_v3.sh. Per-video records and summaries are stored in results\_v3\_gpu1/ and results\_v3\_cpu\_t8/. All primary timing tables use this unified run.

\subsection{GPU Timing Distribution}\label{sup:d.2}

Table~\ref{sup:tab:d1} reports GPU processing time per feature step for the five pipelines under MiniROAD and CMeRT.

\begin{table*}[!htbp]
\centering

\renewcommand{\thetable}{D1}
\caption{GPU time in ms/feature step. Full includes the MiniROAD MV-only auxiliary and routing; TV-L1 denotes RGB+Flow. Standard deviation is across per-video means.}\label{sup:tab:d1}

\small\setlength{\tabcolsep}{3pt}
\begin{lrbox}{\suppTableBox}\begin{tabular}{@{}lrrrrrrr@{}}\toprule
Detector & Pipeline & Mean & p50 & p95 & \shortstack{Across-video std} & \shortstack{Video-mean range} & \shortstack{Decode / prep /\\encoder /\\detector} \\
\midrule
CMeRT & RGB-only & 32.8 & 32.0 & 35.8 & 2.0 & 30.5--38.0 & 3.1 / 1.6 / 7.3 / 20.7 \\
CMeRT & RGB+MV baseline & 58.9 & 57.3 & 64.6 & 4.7 & 52.9--73.2 & 13.4 / 8.7 / 15.0 / 21.8 \\
CMeRT & Full & 59.7 & 58.4 & 64.0 & 3.2 & 54.5--66.4 & 13.3 / 8.7 / 15.0 / 22.7 \\
CMeRT & TV-L1 & 94.4 & 90.3 & 145.2 & 19.0 & 66.3--141.1 & 3.7 / 53.5 / 15.1 / 22.1 \\
CMeRT & RGB+RAFT & 117.8 & 117.4 & 122.0 & 1.5
& 115.3--120.8 & 3.3 / 78.4 / 14.5 / 21.5 \\
MiniROAD & RGB-only & 12.1 & 12.0 & 14.0 & 0.8 & 10.4--13.7 & 3.0 / 1.6 / 6.9 / 0.6 \\
MiniROAD & RGB+MV baseline & 36.2 & 35.7 & 41.0 & 2.2 & 32.4--39.7 & 12.3 / 8.3 / 14.7 / 0.9 \\
MiniROAD & Full & 37.4 & 36.6 & 41.3 & 2.2 & 33.8--43.4 & 12.7 / 8.3 / 14.8 / 1.6 \\
MiniROAD & TV-L1 & 74.5 & 69.2 & 124.8 & 20.0 & 41.9--122.1 & 3.5 / 53.9 / 16.1 / 1.0 \\
MiniROAD & RGB+RAFT & 98.0 & 96.7 & 102.3 & 3.7
& 94.8--112.8 & 3.3 / 78.9 / 14.9 / 0.9 \\
\bottomrule\end{tabular}\end{lrbox}
\ifdim\wd\suppTableBox>\linewidth
  \resizebox{\linewidth}{!}{\usebox{\suppTableBox}}%
\else
  \usebox{\suppTableBox}%
\fi
\end{table*}

\subsection{CPU Timing Distribution}\label{sup:d.3}

Table~\ref{sup:tab:d2} reports CPU processing time per feature step for the five pipelines under MiniROAD and CMeRT.

\begin{table*}[!htbp]
\centering

\renewcommand{\thetable}{D2}
\caption{CPU time in ms/feature step with eight threads. Pipeline definitions and statistics follow Table D1.}\label{sup:tab:d2}

\small\setlength{\tabcolsep}{3pt}
\begin{lrbox}{\suppTableBox}\begin{tabular}{@{}lrrrrrrr@{}}\toprule
Detector & Pipeline & Mean & p50 & p95 & \shortstack{Across-video std} & \shortstack{Video-mean range} & \shortstack{Decode / prep /\\encoder /\\detector} \\
\midrule
CMeRT & RGB-only & 89.1 & 87.0 & 101.8 & 2.7 & 83.7--94.6 & 3.5 / 1.2 / 37.3 / 47.1 \\
CMeRT & RGB+MV baseline & 144.9 & 141.4 & 166.4 & 3.6 & 137.5--151.6 & 13.2 / 6.2 / 74.3 / 51.2 \\
CMeRT & Full & 149.5 & 145.0 & 174.5 & 5.8 & 141.6--163.2 & 12.8 / 6.3 / 76.1 / 54.3 \\
CMeRT & TV-L1 & 1153.2 & 1043.2 & 2235.0 & 418.9 & 572.9--2213.3 & 3.5 / 1011.6 / 84.4 / 53.6 \\
CMeRT & RGB+RAFT & 1244.7 & 1240.7 & 1346.2 & 23.9
& 1185.0--1281.4 & 3.5 / 1114.0 / 73.4 / 53.8 \\
MiniROAD & RGB-only & 45.9 & 44.4 & 56.7 & 1.8 & 42.7--49.6 & 3.6 / 1.3 / 37.9 / 3.2 \\
MiniROAD & RGB+MV baseline & 95.5 & 93.9 & 109.1 & 3.6 & 87.3--101.5 & 12.5 / 6.2 / 74.3 / 2.5 \\
MiniROAD & Full & 98.9 & 96.8 & 113.6 & 3.6 & 90.7--105.1 & 12.9 / 6.2 / 75.0 / 4.7 \\
MiniROAD & TV-L1 & 1100.6 & 990.1 & 2234.1 & 423.2 & 513.2--2155.9 & 3.6 / 1010.8 / 83.8 / 2.4 \\
MiniROAD & RGB+RAFT & 1193.1 & 1189.7 & 1298.3 & 17.3
& 1169.6--1235.9 & 3.6 / 1113.0 / 73.5 / 3.0 \\
\bottomrule\end{tabular}\end{lrbox}
\ifdim\wd\suppTableBox>\linewidth
  \resizebox{\linewidth}{!}{\usebox{\suppTableBox}}%
\else
  \usebox{\suppTableBox}%
\fi
\end{table*}

\subsection{Component Interpretation and Variability}\label{sup:d.4}

Decode includes decoding, MV extraction, and buffering. Prep includes CPU preprocessing and copies, as well as motion estimation for TV-L1 and RAFT. Encoder covers the RGB and motion encoders. Detector includes the main detector, the auxiliary where present, and routing. Rounded components need not sum exactly to rounded totals. The detector-stage increments in Table~\ref{sup:tab:d4} are computed from unrounded measurements.

TV-L1 processing varies substantially across videos because convergence-based early stopping makes its iteration count content-dependent. In the CPU CMeRT flow pipeline, per-video mean prep time ranges from 436 to 2072 ms, with a maximum individual step of approximately 10.3 s. Across-video standard deviations of total latency are 418.9 and 423.2 ms for the two CPU flow pipelines, compared with 3.6--5.8 ms for their MV pipelines. These measurements characterize variability in this implementation and video sample.

Detector-stage increments isolate the auxiliary computation, whereas whole-pipeline differences also reflect variation in decoding and feature encoding. We report both measures to characterize routing overhead.

\subsection{Auxiliary Parameters and Incremental Routing Time}\label{sup:d.5}

The unified MV-only MiniROAD auxiliary shares the upstream motion decoder and feature encoders with the main detector. Table~\ref{sup:tab:d3} compares its size with architecture-matched auxiliary detectors; Table~\ref{sup:tab:d4} isolates its incremental detector-stage latency.

\par\addvspace{4pt}\noindent\begin{minipage}{\columnwidth}
\centering

\renewcommand{\thetable}{D3}
\captionof{table}{Temporal detector parameter counts (millions). Shared feature encoders are excluded.}\label{sup:tab:d3}

\small\setlength{\tabcolsep}{3pt}
\begin{lrbox}{\suppTableBox}\begin{tabular}{@{}lrrr@{}}\toprule
\shortstack{Main detector} & RGB+MV main & \shortstack{Same-architecture\\MV-only auxiliary} & \shortstack{Unified MiniROAD\\MV-only auxiliary} \\
\midrule
MiniROAD & 17.86M & 13.67M & 13.67M \\
CMeRT & 96.78M & 88.38M & 13.67M \\
LSTR & 60.98M & 56.78M & 13.67M \\
TeSTra & 60.98M & 56.78M & 13.67M \\
\bottomrule\end{tabular}\end{lrbox}
\ifdim\wd\suppTableBox>\linewidth
  \resizebox{\linewidth}{!}{\usebox{\suppTableBox}}%
\else
  \usebox{\suppTableBox}%
\fi
\end{minipage}\par\addvspace{4pt}

The shared auxiliary contains 13.67M parameters, reducing the auxiliary parameter count by 84.5\% for CMeRT and 75.9\% for LSTR and TeSTra relative to their architecture-matched auxiliaries.

\par\addvspace{4pt}\noindent\begin{minipage}{\columnwidth}
\centering

\renewcommand{\thetable}{D4}
\captionof{table}{Incremental detector-stage latency for the shared MiniROAD MV-only auxiliary (ms/feature step; computed before rounding).}\label{sup:tab:d4}

\small\setlength{\tabcolsep}{3pt}
\begin{lrbox}{\suppTableBox}\begin{tabular}{@{}lrr@{}}\toprule
\shortstack{Main detector} & \shortstack{GPU increment} & \shortstack{CPU increment} \\
\midrule
MiniROAD & 0.76 & 2.26 \\
CMeRT & 0.92 & 3.14 \\
\bottomrule\end{tabular}\end{lrbox}
\ifdim\wd\suppTableBox>\linewidth
  \resizebox{\linewidth}{!}{\usebox{\suppTableBox}}%
\else
  \usebox{\suppTableBox}%
\fi
\end{minipage}\par\addvspace{4pt}

Subtracting the displayed GPU means in main-text Table 4 gives 1.2 ms/step for MiniROAD and 0.8 ms/step for CMeRT. Computed before rounding, the whole-pipeline increments are 1.27/3.32 ms for MiniROAD and 0.80/4.61 ms for CMeRT on GPU/CPU, respectively.
\clearpage

\section*{Supplementary References}

\textbf{{[}S1{]}} Zachary Teed, Jia Deng. ``RAFT: Recurrent All-Pairs Field Transforms for Optical Flow.'' In \emph{Computer Vision -- ECCV 2020}, pp.~402--419, 2020. DOI: \url{https://doi.org/10.1007/978-3-030-58536-5_24}

\textbf{{[}S2{]}} Thomas Wiegand, Gary J. Sullivan, Gisle Bj{\o}ntegaard, Ajay Luthra. ``Overview of the H.264/AVC Video Coding Standard.'' \emph{IEEE Transactions on Circuits and Systems for Video Technology}, vol.~13, no. 7, pp.~560--576, 2003. DOI: \url{https://doi.org/10.1109/TCSVT.2003.815165}

\textbf{{[}S3{]}} NVIDIA. \href{https://docs.nvidia.com/video-technologies/pynvvideocodec/pynvc-api-prog-guide/using_pynvvideocodec_apis.html}{\emph{Using PyNvVideoCodec APIs}}. NVIDIA PyNvVideoCodec API Programming Guide, section ``Latency Modes.''  Accessed 2026-10-12.

\textbf{{[}S4{]}} Amazon Web Services. \href{https://docs.aws.amazon.com/ivs/latest/RealTimeUserGuide/rt-stream-ingest.html}{\emph{IVS Stream Ingest: Real-Time Streaming}}. Amazon IVS Real-Time Streaming User Guide. 2026.  Accessed 2026-10-12.


\end{document}